\documentclass{article}

\usepackage[preprint, nonatbib]{neurips_2023}

\usepackage[utf8]{inputenc} 
\usepackage[T1]{fontenc}    
\usepackage{hyperref}       
\usepackage{url}            
\usepackage{booktabs}       
\usepackage{amsfonts}       
\usepackage{nicefrac}       
\usepackage{microtype}      
\usepackage{xcolor}         

\usepackage{caption}
\usepackage{subcaption}
\usepackage{graphicx}
\usepackage{float}
\usepackage{multirow}
\usepackage{tabularx}
\usepackage{arydshln}
\usepackage{wrapfig}

\usepackage{algorithm}
\usepackage{algpseudocode}
\usepackage{physics,amsmath,amssymb}
\usepackage{multirow}

\usepackage{color}
\usepackage{float}

\graphicspath{{figs/}}

\usepackage[compact]{titlesec}
\titlespacing{\section}{0px}{1em}{0px}
\titlespacing{\subsection}{0px}{1em}{0px}
\titlespacing{\subsubsection}{0px}{0px}{0px}
\titlespacing{\paragraph}{0px}{0px}{0.5em}

\title{GPT Self-Supervision for a Better Data Annotator}
\author{%
  Xiaohuan Pei, Yanxi Li,  Chang Xu \\
  School of Computer Science, Faculty of Engineering\\
  The University of Sydney\\
  \texttt{xpei8318@uni.sydney.edu.au, yali0722@uni.sydney.edu.au, c.xu@sydney.edu.au} \\
}

\begin{document}

\maketitle

\begin{abstract}
The task of annotating data
into concise summaries poses a significant challenge across various domains, frequently requiring the allocation of significant time and specialized knowledge by human experts. 
Despite existing efforts to use large language models for annotation tasks, significant problems such as limited applicability to unlabeled data, the absence of self-supervised methods, and the lack of focus on complex structured data still persist.
In this work, we propose a GPT self-supervision annotation method, which embodies a generating-recovering paradigm that leverages the one-shot learning capabilities of the Generative Pretrained Transformer (GPT).
The proposed approach comprises a one-shot tuning phase followed by a generation phase. In the one-shot tuning phase, 
we sample a data from the support set as part of the prompt for GPT to generate a textual summary, which is then used to recover the original data. The alignment score between the recovered and original data serves as a self-supervision navigator to refine the process. 
In the generation stage, the optimally selected one-shot sample serves as a template in the prompt and is applied to generating summaries from challenging datasets. 
The annotation performance is evaluated by tuning several human feedback reward networks and by calculating alignment scores between original and recovered data at both sentence and structure levels. Our self-supervised annotation method consistently achieves competitive scores, convincingly demonstrating its robust strength in various data-to-summary annotation tasks.
\end{abstract}

\section{Introduction}
Large language models represented by Generative Pre-trained Transformer(GPT) family \cite{radford2018improving, radford2019language, brown2020language} have made breakthrough advancements in recent years, achieving state-of-the-art performance across various deep learning tasks. Among these tasks, data annotation is the fundamental and indispensable first step in the process of AI research, as it lays the groundwork by providing labeled data that serves as the foundation for training and evaluating models \cite{laga2020survey, yang2023learning, cai2023semi, chen2023smg}. The quality of data label generation directly impacts all subsequent tasks, making it the cornerstone of the entire deep learning process \cite{kopp2023tackling, parikh2020totto}. 

The annotation task poses the significant challenges. First, the complexity of the data makes it time-consuming for experts to annotate each sample \cite{mcever2023context, yang2023dynamic, Chen2021, Zhang2023}. 
For example, computational graphs provide valuable insights into the design of neural network structures and annotating these cells requires significant effort to identify and summarize similar sub-sequences as parallel blocks \cite{dobre2022immersive}, which is nearly impossible to identify and isolate.
Second, the exclusive use of subjective human-annotated summaries presents evaluation challenges \cite{deriu2021survey, brown2020language}. Employing these subjective summaries as supervised learning labels could lead to annotation tasks reflecting human bias, not the impartial correlation between the data and its summary.

Recent research has harnessed the capabilities of large language models for supervised data annotation. As revealed by \cite{ding2022gpt, gilardi2023chatgpt, zhu2023can}, GPT models can annotate classification tasks through the direct design of diverse prompts. Furthermore, the study by \cite{labruna2023unraveling} broadens the scope of this annotation task, enabling the generation of dialogues.
Despite impressive progress, the problems of annotation task still remain. 
First, current studies are primarily centered around supervised annotation with human-labeled data, which restricts the annotation method's applicability to more general unlabeled data. In most scenarios, human-labeled data is unavailable to guide the annotation process. 
Second, prior research has been limited to the design of various prompts, without considering feedback mechanisms to refine specific parts of the prompts. 
Third, there's a lack of research focused on annotating complex structured data. The current work only focuses on annotating simple natural language description samples without challenging more complex structured data. For example, the datasets of computational graphs extracted by neural networks consists various nested structure, making it difficult for humans to annotate specific blocks within such complex lists of edges.

\begin{wrapfigure}{r}{0.5\textwidth}
  \vspace{-0.5em}
  \setlength{\abovecaptionskip}{0.00cm}
  \setlength{\belowcaptionskip}{0.00cm}
  \begin{center}
    \includegraphics[width=0.49\textwidth]{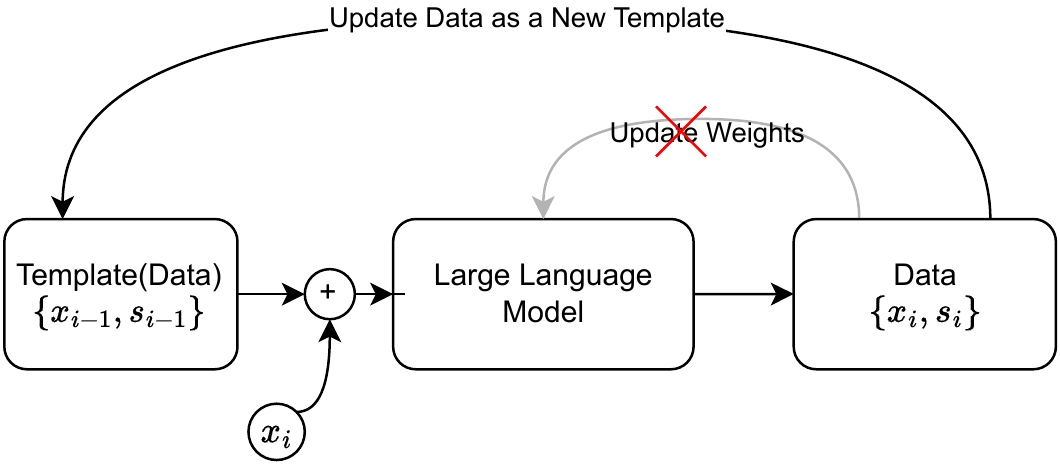}
  \end{center}
  \caption{Data-centric based generating-recovering paradim. Iterative update the template with the generated data. $x_i$ is sampled from a candidate support set and $s_i$ is generated by GPT. }
  \vspace{-0.2in}
  \label{fig:1}
\end{wrapfigure}
Here we propose a GPT self-supervision approach via a generating-recovering loop, primarily inspired by prompt tuning \cite{lester2021power, lester2022guiding} and the data-centric paradigm \cite{polyzotis2021datacentric}, as outlined in Figure \ref{fig:1}.
The framework of self-supervision method contains the one-shot phase and the generation phase. 
The one-shot phase aims to iteratively find the optimal pairs of the data and summary as the template. 
Starting with a human-annotated data pair, our iterative process utilizes GPT to generate summaries. The generated summary and raw data form a new data pair. This new data pair is then evaluated for its potential as a template through a comparison of the recovered data from the summary with the original. If both the support and validation scores improve, we update the optimal template with the current new data pair.
Thus, the self-alignment mechanism iteratively tune the one-shot template for the next round generation.
The searched optimal template is subsequently used for generating the summaries on the generation phase. We tune various reward models to evaluate summary quality and introduce various similarity metrics to assess the recovery ability. We perform sufficient experiments on three challenging datasets and conduct detailed ablation study from various perspectives. The results demonstrate our self-supervision paradigm consistently yields competitive performance evaluated by reward models and recovery scores. Addtionally, we apply our self-supervision method to generate two new datasets of 3k/15k summaries of the neural network architectures based on the different computational operators.

\section{Related Work}
    
    \paragraph{Large Language Model.}
    GPT-1 \cite{radford2018improving} propose a two-step approach: pre-training on unlabeled text and discriminative fine-tuning on specific tasks. Unlike previous methods, It employes task-aware input transformations during fine-tuning, minimizing changes to the model architecture.
    GPT-2 \cite{radford2019language} demonstrates language models can perform down-stream tasks in a zero-shot setting – without any parameter or architecture modification.
    GPT-3 \cite{brown2020language} scales up language models to improve their ability with minimal fine-tuning abn achieves strong few-shot performance without gradient updates or task-specific fine-tuning.

    \paragraph{One/Few-Shot Learning.}
    Learning new concepts quickly with limited data is a challenge in machine learning. Traditional supervised deep learning is not effective for this. 
    Li \textit{et al.} \cite{fei2006one} use Bayesian modeling and show that it's possible to learn a lot about a category from just one or a few images by leveraging knowledge from previously learned categories, regardless of their differences.
    Vinyals \textit{et al.} \cite{vinyals2016matching} define one-shot learning problems and combine metric learning and neural networks with external memories to create a framework that can learn new concepts from just a few examples. The framework doesn't require fine-tuning and is tested on vision and language tasks.
    As aforementioned, GPT-3 \cite{brown2020language} is a powerful few-shot learner, exhibiting remarkable performance on various natural language processing tasks, including translation, question-answering, and cloze tasks, without the need for fine-tuning or gradient updates.

\section{Approach}
\begin{figure}[!tbp]
    \centering
    \includegraphics[width=\linewidth]{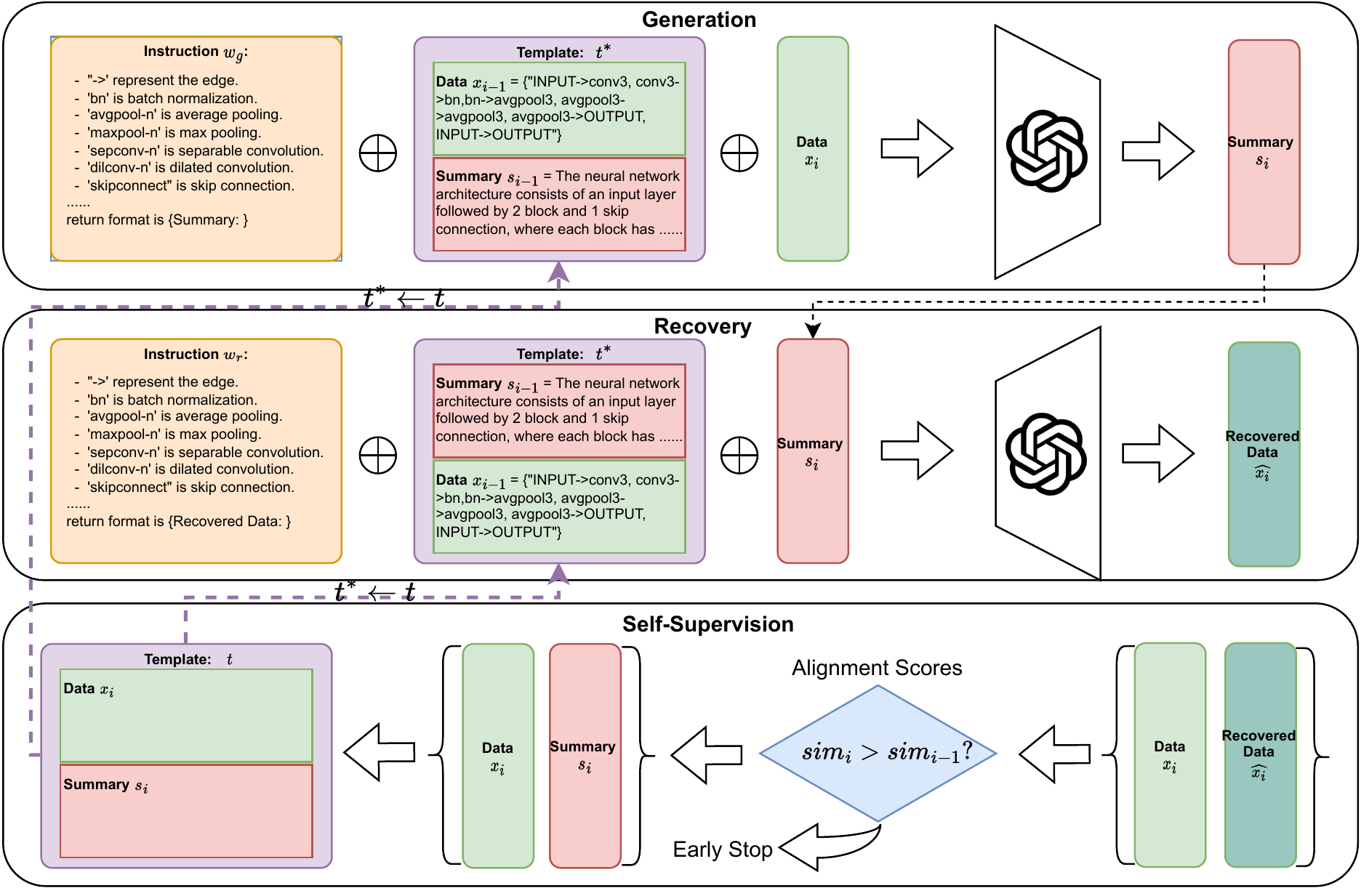}
    \caption{GPT Self-Supervision Annotation: A framework of the generating-recovering paradigm, where the objective function seeks to maximize the alignment scores between the original data and the recovered data.}
    \label{fig:annotation}
\end{figure}
Our data annotation approach involves \textbf{one-shot tuning stage} and \textbf{summary generation stage}.

\paragraph{One-shot tuning stage.}
This stage is mainly responsible for finding the optimal template $t^{*}$ self-supervised by GPT.
It is an iterative process of generating a summary, recovering data, and comparing feedback values to navigate the template tuning in the prompt, which is subsequently used for the next round generation.

The iterative process contains the following steps. 
We first initialize a simple human-labeled pairs of data $x_0$, summary $s_0$ as the best template $t^{*} = \{x_0, s_0\}$, and then respectively assign the role type of \{system, assistant, content\} for the instruction, template, support data $\{w, t, x\}$.
For iteration $i$, we sample a supported data $x_i$ from the support set $\mathcal{X}$, which are subsequently concatenated with the current optimal template $t^{*}$ and default instruction $w_g$ into one message.  The GPT $\mathcal{F}(\cdot)$ covert the message to generate summary $s_i$ by referring the instruction, current template and data:
\begin{equation}
    s_i \leftarrow \mathcal{F}(x_i|t_i, w_g, \theta),
    \label{eq:1}
\end{equation}
where the $\theta$ is the parameters of the language model function and $w_g$ is the instruction of generating a summary. At this point, we obtain a new paired set consisting of data and summary, denoted as  $\{x_i, s_i\}$, where $s_i$ is generated from $x_i$.

Considering that a summary's main purpose is to briefly capture the essence of a dataset, the quality of a summary naturally  can be deduced from its ability to faithfully reproduce the original dataset.
The recovering process is reconstructing $\hat{x}_i$ from $s_i$ by the same GPT:
\begin{equation}
    \hat{x}_i \leftarrow \mathcal{F}(s_i|t_i, w_r, \theta),
    \label{eq:2}
\end{equation}
where the $\theta$ is the parameters of the large language model and $w_g$ is the instruction of recovering data. And then we apply $sim(\hat{x}_i, x_i)$ to measure the similarity score between the recovered data and original data. If the current score surpasses the previously recorded highest value from iterations, we consider the current data-summary pairs \{$x_i$, $s_i$\} as a temporary template.

Using the same generation and recovery process mentioned in \ref{eq:1} \ref{eq:2}, we evaluate the average similarity score on the validation set. If this score remains higher than the maximum valid score observed in previous iterations, the self-supervised mechanism update the best template $t^*$ with the current data-summary pairs \{$x_i$, $s_i$\}.
\begin{algorithm}[t]
  \setlength{\abovecaptionskip}{0.00cm}
  \setlength{\belowcaptionskip}{0.00cm}
\caption{Self-Supervised Annotating by GPT.
}
\begin{algorithmic}[1]
\State \textbf{Initialize:} instructions $w_g$/$w_r$ for generating/recovering , optimal one-shot template $t^*=\{x_0, s_0\}$, best support/valid similarity score $sim^{sup}/sim^{val} = 0$, a support set $\mathcal{X}^{sup}$ for sampling one-shot template, a validation set $\mathcal{X}^{val}$ for testing the new template performance.
\For{iteration $i \gets 1$ to $I$}
    \State \textbf{(1) Sampling:} $x_i \leftarrow \mathcal{X}^{sup}$
    \State \textbf{(2) Encoding:} Send the message <\textit{instruction, template, sampled data}> to GPT:
    \State $$s_i \leftarrow GPT(\langle w_g, t^*, x_i \rangle)$$
    \State Get the response of summary $s_i$
    \State \textbf{(3) Decoding:} Send the message <\textit{instruction, template, generated summary}> to GPT:    
    \State $$\hat{x}_i \leftarrow GPT(\langle w_r, t^*, s_i \rangle)$$
    \State Get the response of recovered data $\hat{x}_i$.
    \State Compute similarity score between recovered data and original data
    \State $$sim^{sup}_i \leftarrow sim(x_i, \hat{x}_i)$$
    \State \textbf{(4) Update:} 
    \If{$sim^{sup}_i > sim^{sup}$}
        \State Compute the average similarity score $sim^{val}_i$ on the validation set $\mathcal{X}^{val}$ by (2)(3).
        \If {$sim^{val}_i > sim^{val}$}
            \State Replace the best similarity scores with current score: 
            \begin{equation*}
                sim^{sup} \leftarrow sim^{sup}_i \text{;} \quad sim^{val} \leftarrow sim^{val}_i
            \end{equation*}
            \State Update the optimal template with support data and generated summary: 
            $$t^{*} \leftarrow \{x_i, s_i\}$$
        \EndIf
    \EndIf
\EndFor
\State \textbf{Output:} Return the best template $t^*$ for the generation phase.
\end{algorithmic}
\end{algorithm}
The process continues until the current sample cannot achieve a higher recovery score compared to the previous iteration or when the maximum number of iterations is achieved. 
And the self-supervised objective of above iterative updating is to find the best template $t_i$ that maximizes the expected similarity between the recovered data and the original data. The optimization problem can be formalized as follows:
\begin{equation}
t^* = \arg\max_{t} \ \mathbb{E}_{{x_i \sim \mathcal{X}}} [sim (\hat{x}_i, x_i)],
\end{equation}
where the $sim(\cdot)$ is the metric of similarity function between two sequence data. 
This objective assumes that the similarity $sim$ is a meaningful measure of the quality of the recovery process, and that higher values of $sim$ correspond to better recoveries. 
And this objective has a few notable elements than traditional one-shot learning stage.
Unlike conventional self-supervised method where model weights are updated during training, our technique updates the current template instead. 

\paragraph{Summary generation stage.}
In this stage, GPT concatenates the identified optimal template with instructions, using it as the optimal prompt to generate natural language summaries for the generation dataset:
\begin{equation}
    s \leftarrow \mathcal{F}(x|t^{*}, w_g, \theta),
\end{equation}
where the $s$ is the summarises of dataset $x$. 
We test the summary quality by tuning various human feedback reward networks, and then we introduce the recovery evaluation aiming to measure whether the summary could decoding the high-level sentences to original data. Specifically, we assume that the stronger the ability to recover the intermediate summary back to the original data, the more it substantiates the professionalism and accuracy of the summarises. Conversely, a weaker recovery capability implies a lower degree of professionalism and accuracy in the summarises. 

The evaluation phase of this process is two-fold. Firstly, we assess the quality of the summary by employing a variety of human feedback reward networks which offers us a comprehensive insight into the real-world applicability and understandability of our summaries. Following this, we institute a recovery evaluation process that seeks to measure the efficiency and accuracy with which the summary can decode high-level sentences back into the original data, thereby serving as an indicator of the professionalism and accuracy of the summarises. This assumption underpins our belief that the more capable the model is of reverse engineering the summary back to the original data, the more it affirms the precision and professionalism of the summaries. Conversely, if the model demonstrates a lower proficiency in this restoration task, it suggests that the summaries lack a certain degree of professionalism and accuracy.

\section{Experiments}
\subsection{Experiment Setup}
    \paragraph{Dataset}
    This study utilizes three distinct datasets: Darts-Medium, Darts-Large, and PubMed. The Darts-Medium and Darts-Large datasets consist of neural architecture networks, generated by 5 and 7 operators respectively, with Darts-Large having more nodes. They provide a rich source of information on the design and performance of various neural architectures. We have also incorporated the PubMed dataset which consists of Isomeric SMILES structures into our study. This dataset represents a different domain, offering an opportunity to examine the generation of high-level summaries for complex chemical structures. We take great care to avoid test set leakage and split data for each stage to use.
Each of these datasets has been divided into two segments for different stages of our experiment: the one-shot phase and the generation phase. The one-shot phase involves a support set and a validation set, each containing $50$ samples. These sets are utilized for finding optimal template from the support set. The generation stage splits the data using a $K$-Fold ($K=5$) method into training and testing sets. The training set is used to further train the model and adjust the reward mechanism, while the test set is reserved for evaluating the summary quality.

    \paragraph{Setup}
    We implement a temperature hyperparameter of 1 to promote diversity in the summary generation when evaluating baseline performance. Conversely, during the data generation stage, the temperature was set to zero to ensure consistency and stability. 
    To test the performance of different large language models, we apply four widely used models in this experiment, composed of the version 3 of GPT: davinci, text-curie-001  and version 3.5 of GPT: text-davinci-003, gpt-3.5-turbo. We call the response through the inference of the openAI official APIs \footnote{\href{https://platform.openai.com/docs/models/gpt-3}{platform.openai.com/docs/models/}}. The initialized information is divided into instructions, templates and query data, respectively calling the role of system, assistant and user in the information flow. We set $10$ iterations in each one-shot tuning phase with a defined maximum token length of $350$ for generating summaries and $500$ for recovering datas from the summaries. Each input prompt consists of three parts, an instruction, a template and a sampled data. We give the input and output formats in the instruction, and define the meaning of each structured symbol in the data. This part costs 500 tokens. For the template, we assign assistant and content tags to the data and summary in the template, so that GPT can recognize that this is a template information, and this part costs 3000 tokens. For sampling data from support set in the one-shot phase, we directly use the role of user to send it to GPT. In order to make the generated summary more diverse in the one-shot tuning stage, we set the hyperparameter of temperature to 1. In the data generation stage, we set the temperature to 0 to keep it stable.

\subsection{Evaluation}\label{Evaluation}
\textbf{Summary Evaluation.} 
To directly evaluate summary quality generated by our approach, we tuned various human feedback reward models as our evaluators. These reward models have been widely acknowledged in the literature for their effectiveness in providing evaluative feedback for language generation tasks
 
Specifically, we tuned various human feedback reward models \footnote{\href{https://huggingface.co/sugam11/gpt2-rlhf-reward}{gpt2-rlhf-reward}}\footnote{\href{https://huggingface.co/OpenAssistant/reward-model-deberta-v3-large}{reward-model-deberta-v3-large}}\footnote{\href{https://huggingface.co/OpenAssistant/reward-model-deberta-v3-large}{reward-model-deberta-v3-base}}\footnote{\href{https://huggingface.co/AdamG012/chat-opt-350m-reward-deepspeed}{chat-opt-350m-reward-deepspeed}} as our evaluators. Each evaluator evaluate summary quality and provide a corresponding reward score ($R1, R2, R3, R4$). To ensure stability in the reward distribution, we employed a k-fold cross-validation strategy with a value of k set to 5. Assume $r(x, s|\theta)$ represents the scalar output of the reward model for data $x$ and summary $s$, parameterized by $\theta$. We performed fine-tuning on each pre-trained reward network by following the steps outlined in \cite{rlhf1}:
\begin{equation}
    \mathcal{L} = \mathbb{E}_{(x, s_0, s_1, s_i)\sim \mathcal{D}}[log(\sigma(r(x, s_i| \theta)) - \sigma(r(x, s_{1-i}|\theta))],
\end{equation}
where $i \in \{0, 1\}$ and $\mathcal{D}$ represent the human-labelled datasets containing judgments on which summary, generated by two large language models, is superior. The equation delineated above illustrates the loss function we used during this fine-tuning process. The reward process symbolize the human-annotated datasets that provide judgments about the superiority of summaries generated by two large language models. This dual evaluation not only adds a level of redundancy but also ensures a more rigorous and comprehensive assessment of the generated summaries.

\textbf{Recovery Evaluation.}
We implement both sentence-level alignments and embedding-level metrics to assess the discrepancy between the recovered data and the original data. 

For the sentence-level evaluation, we employed Average BLEU score\cite{bleu} and ROUGE-L\cite{rouge1, rouge2}. 
Specifically, we utilize a smoothed average version of BLEU in our evaluations to counteract the issues that can arise with BLEU when dealing with short sentences. ROUGE-L, on the other hand, is based on Longest Common Subsequence (LCS) statistics, which makes it a robust tool for evaluating the quality of summaries, particularly in our case where it was applied for the evaluations. 

When it comes to embedding-level metrics, we employ the Semantic Textual Similarity (STS) \cite{mikolov2013efficient} embedding and Bidirectional Encoder Representations from Transformers (BERT) \cite{devlin2018bert} embedding. The STS embedding, in particular, offers a quantifiable measure of the semantic equivalence between two text pieces, which is ideal for assessing the semantic similarity between the original and recovered data. On the other hand, the BERT embedding, which originates from the BERT model, allows us to capture more nuanced semantic and syntactic features of the data structures, providing a more comprehensive and insightful analysis of our generated summaries compared to the original data.

\begin{table}[!tbp] 
\vspace{-0.1cm}
\caption{Various evaluation scores ($\pm $ stardard error) on the three datasets. The performance of each model is evaluated using two types of evaluative metrics.}
\begin{center}
\resizebox{\textwidth}{!}{
\begin{tabular}{lccccccccc}
\hline
\multicolumn{1}{c}{\multirow{2}{*}{\textbf{Model}}} & \multicolumn{4}{c}{\textbf{Generated Summary Evaluation}} & \multicolumn{4}{c}{\textbf{Recovered Data Evaluation}} \\
\cline{2-5} \cline{6-9}
 & \textbf{R1} & \textbf{R2} & \textbf{R3} & \textbf{R4} & \textbf{BLEU} & \textbf{ROUGE} & \textbf{STS Sim} & \textbf{Bert Sim} \\
\hline
& \multicolumn{8}{c}{Darts-Medium} \\
\hdashline
davinci & $0.296_{\small{\pm 0.012}}$ & $0.302_{\small{\pm 0.041}}$ & $0.194_{\small{\pm 0.016}}$ & $0.288_{\small{\pm 0.019}}$ & $0.195_{\small{\pm 0.004}}$ & $0.214_{\small{\pm 0.007}}$ & $0.291_{\small{\pm 0.001}}$ & $0.263_{\small{\pm 0.005}}$ \\
text-curie-001 & $0.243_{\small{\pm 0.028}}$ & $0.211_{\small{\pm 0.013}}$ & $0.297_{\small{\pm 0.004}}$ & $0.342_{\small{\pm 0.013}}$ & $0.118_{\small{\pm 0.004}}$ & $0.172_{\small{\pm 0.008}}$ & $0.310_{\small{\pm 0.006}}$ & $0.294_{\small{\pm 0.009}}$ \\
text-davinci-003 & $0.532_{\small{\pm 0.023}}$ & $0.596_{\small{\pm 0.028}}$ & $0.503_{\small{\pm 0.032}}$ & $0.582_{\small{\pm 0.031}}$ & $0.278_{\small{\pm 0.006}}$ & $\textbf{0.379}_{\small{\pm 0.017}}$ & $0.772_{\small{\pm 0.003}}$ & $0.543_{\small{\pm 0.025}}$ \\
gpt-3.5-turbo & $\textbf{0.513}_{\small{\pm 0.011}}$ & $\textbf{0.642}_{\small{\pm 0.016}}$  & $\textbf{0.519}_{\small{\pm 0.015}}$ & $\textbf{0.639}_{\small{\pm 0.017}}$ & $\textbf{0.482}_{\small{\pm 0.023}}$ & $0.422_{\small{\pm 0.004}}$ & $\textbf{0.816}_{\small{\pm 0.014}}$ & $\textbf{0.691}_{\small{\pm 0.002}}$ \\ 
\hdashline
& \multicolumn{8}{c}{Darts-Large} \\
\hdashline
davinci & $0.302_{\small{\pm 0.024}}$ & $0.294_{\small{\pm 0.053}}$ & $0.197_{\small{\pm 0.023}}$ & $0.305_{\small{\pm 0.025}}$ & $0.194_{\small{\pm 0.012}}$ & $0.221_{\small{\pm 0.017}}$ & $0.287_{\small{\pm 0.009}}$ & $0.268_{\small{\pm 0.045}}$ \\
text-curie-001 & $0.252_{\small{\pm 0.038}}$ & $0.220_{\small{\pm 0.023}}$ & $0.305_{\small{\pm 0.011}}$ & $0.370_{\small{\pm 0.021}}$ & $0.129_{\small{\pm 0.014}}$ & $0.182_{\small{\pm 0.022}}$ & $0.314_{\small{\pm 0.013}}$ & $0.308_{\small{\pm 0.019}}$ \\
text-davinci-003 & $0.509_{\small{\pm 0.034}}$ & $0.607_{\small{\pm 0.021}}$ & $0.515_{\small{\pm 0.011}}$ & $0.580_{\small{\pm 0.027}}$ & $0.292_{\small{\pm 0.005}}$ & $0.382_{\small{\pm 0.013}}$ & $0.781_{\small{\pm 0.001}}$ & $0.559_{\small{\pm 0.009}}$ \\
gpt-3.5-turbo & $\textbf{0.544}_{\small{\pm 0.020}}$ & $\textbf{0.672}_{\small{\pm 0.037}}$ & $\textbf{0.537}_{\small{\pm 0.021}}$ & $\textbf{0.657}_{\small{\pm 0.033}}$ & $\textbf{0.505}_{\small{\pm 0.030}}$ & $\textbf{0.447}_{\small{\pm 0.009}}$ & $\textbf{0.829}_{\small{\pm 0.019}}$ & $\textbf{0.715}_{\small{\pm 0.006}}$ \\
\hdashline
& \multicolumn{8}{c}{PubMed} \\
\hdashline
davinci & $0.318_{\small{\pm 0.017}}$ & $0.354_{\small{\pm 0.026}}$ & $0.216_{\small{\pm 0.032}}$ & $0.310_{\small{\pm 0.031}}$ & $0.209_{\small{\pm 0.015}}$ & $0.228_{\small{\pm 0.002}}$ & $0.293_{\small{\pm 0.002}}$ & $0.265_{\small{\pm 0.002}}$ \\
text-curie-001 & $0.262_{\small{\pm 0.013}}$ & $0.230_{\small{\pm 0.021}}$ & $0.327_{\small{\pm 0.018}}$ & $0.372_{\small{\pm 0.026}}$ & $0.138_{\small{\pm 0.012}}$ & $0.192_{\small{\pm 0.001}}$ & $0.316_{\small{\pm 0.001}}$ & $0.310_{\small{\pm 0.001}}$ \\
text-davinci-003 & $\textbf{0.547}_{\small{\pm 0.019}}$ & $0.611_{\small{\pm 0.022}}$ & $0.514_{\small{\pm 0.036}}$ & $0.594_{\small{\pm 0.023}}$ & $\textbf{0.571}_{\small{\pm 0.008}}$ & $0.403_{\small{\pm 0.001}}$ & $0.774_{\small{\pm 0.004}}$ & $0.571_{\small{\pm 0.002}}$ \\
gpt-3.5-turbo & $0.542_{\small{\pm 0.022}}$ & $\textbf{0.671}_{\small{\pm 0.031}}$ & $\textbf{0.547}_{\small{\pm 0.023}}$ & $\textbf{0.667}_{\small{\pm 0.012}}$ & $0.509_{\small{\pm 0.004}}$ & $\textbf{0.457}_{\small{\pm 0.001}}$ & $\textbf{0.796}_{\small{\pm 0.001}}$ & $\textbf{0.635}_{\small{\pm 0.002}}$ \\
\hline
\end{tabular}}
\label{tab:results}
\vspace{-0.4cm}
\end{center}
\end{table}

\subsection{Baseline Results}
We conducted experiments on three highly challenging datasets using four different models: davinci, text-curie-001, text-davinci-003, gpt-3.5-turbo. The evaluation included testing the summary quality by four specific reward networks and assessing the data recovery capability by alignment scores on both sentence and embedding levels.

\paragraph{Evaluate on the Recovery Data}
We first focus on the evaluation of the generated summary using the four reward scores. These scores represent the performance of the generated summary quality by various human feedback reward models after fine-tuning. On the Darts-Medium dataset, GPT-3.5-Turbo outperforms the other models with the highest scores across all metrics: a BLEU score of $0.482 \pm 0.023$, a ROUGE score of $0.422 \pm 0.004$, a STS Sim score of $0.816 \pm 0.014$, and a Bert Sim score of $0.691 \pm 0.002$. The Text-davinci-003 model followed next, while the Davinci and Text-curie-001 models lag behind, with lower scores on all measures. Similar trends were observed on the Darts-Large dataset. Once again, GPT-3.5-Turbo displays superior performance, achieving the highest scores in all categories: BLEU ($0.505 \pm 0.030$), ROUGE ($0.447 \pm 0.009$), STS Sim ($0.829 \pm 0.019$), and Bert Sim ($0.715 \pm 0.006$). Text-davinci-003 maintains its second place ranking, while Davinci and Text-curie-001 trails with less impressive scores. On the PubMed dataset, the Text-davinci-003 model remarkably achieved the highest BLEU score of $0.571 \pm 0.008$, surpassing the other models. However, GPT-3.5-Turbo still led the other metrics, with a ROUGE score of $0.457 \pm 0.001$, a STS Sim score of $0.796 \pm 0.001$, and a Bert Sim score of $0.635 \pm 0.002$. Davinci and Text-curie-001 continued to exhibit inferior performance compared to the other models. While the performance varied somewhat across different datasets, GPT-3.5-Turbo consistently yields the strongest results on the evaluated metrics. This strongly indicates its superior capability in data recovery tasks. The performance gap observed between the models highlights the importance of selecting the right transformer model for specific tasks and datasets, thereby optimizing the trade-off between computational resources and performance.

\paragraph{Evaluate on the Summary Quality}
The results in Table \ref{tab:results} demonstrate that on the Darts-Medium dataset, Text-davinci-003 displays the best overall performance in terms of R1 ($0.532 \pm 0.023$), R2 ($0.596 \pm 0.028$), and R3 ($0.503 \pm 0.032$) scores. However, GPT-3.5-Turbo achieves the highest R4 score ($0.639 \pm 0.017$). The other models, Davinci and Text-curie-001, have lower scores across these metrics.
On the Darts-Large dataset, GPT-3.5-Turbo outperformed the other models across all reward metrics, with R1 ($0.544 \pm 0.020$), R2 ($0.672 \pm 0.037$), R3 ($0.537 \pm 0.021$), and R4 ($0.657 \pm 0.033$) scores. Text-davinci-003 follows closely behind, while Davinci and Text-curie-001 lag further in terms of performance.
For the PubMed dataset, Text-davinci-003 demonstrate superior performance in the R1 ($0.547 \pm 0.019$), R2 ($0.611 \pm 0.022$), and R3 ($0.514 \pm 0.036$) metrics, while GPT-3.5-Turbo achieved the highest R4 score ($0.667 \pm 0.012$). As in the previous datasets, Davinci and Text-curie-001 exhibit lower scores across these reward metrics.
Based on the performance, the evaluation on the reward models reveals that Text-davinci-003 and GPT-3.5-Turbo consistently outperform the other models in terms of R1, R2, R3, and R4 scores. These findings emphasize the importance of selecting the appropriate reward models for specific tasks and datasets, as they have a significant impact on the quality of the generated summaries.

\paragraph{Comparison of Datasets}
Simultaneously, we observe that the annotation approach yields varying results across different datasets. Firstly, it's obvious that the complexity and characteristics of the dataset have a significant impact on the performance of the models. For instance, on the Darts-Medium dataset, while the GPT-3.5-Turbo model performs exceptionally well on the data recovery metrics (BLEU, ROUGE, STS Sim, and Bert Sim), its performance in terms of the reward models (R1, R2, R3, and R4) was surpassed by the Text-davinci-003 model. However, the scenario was slightly different on the Darts-Large dataset, where GPT-3.5-Turbo outperforms the other models in all evaluated metrics. This implies that the model was more adept at handling the increased complexity and volume of this dataset. 
On the other hand, the models' performance on the PubMed dataset presents a more balanced records, where the Text-davinci-003 model surpasses others in terms of the BLEU score and the R1, R2, and R3 reward model scores, while GPT-3.5-Turbo leads in the other metrics. 

\begin{figure}[!tbp]
    \centering
    \begin{subfigure}{0.245\textwidth}
        \centering
        \includegraphics[width=\textwidth]{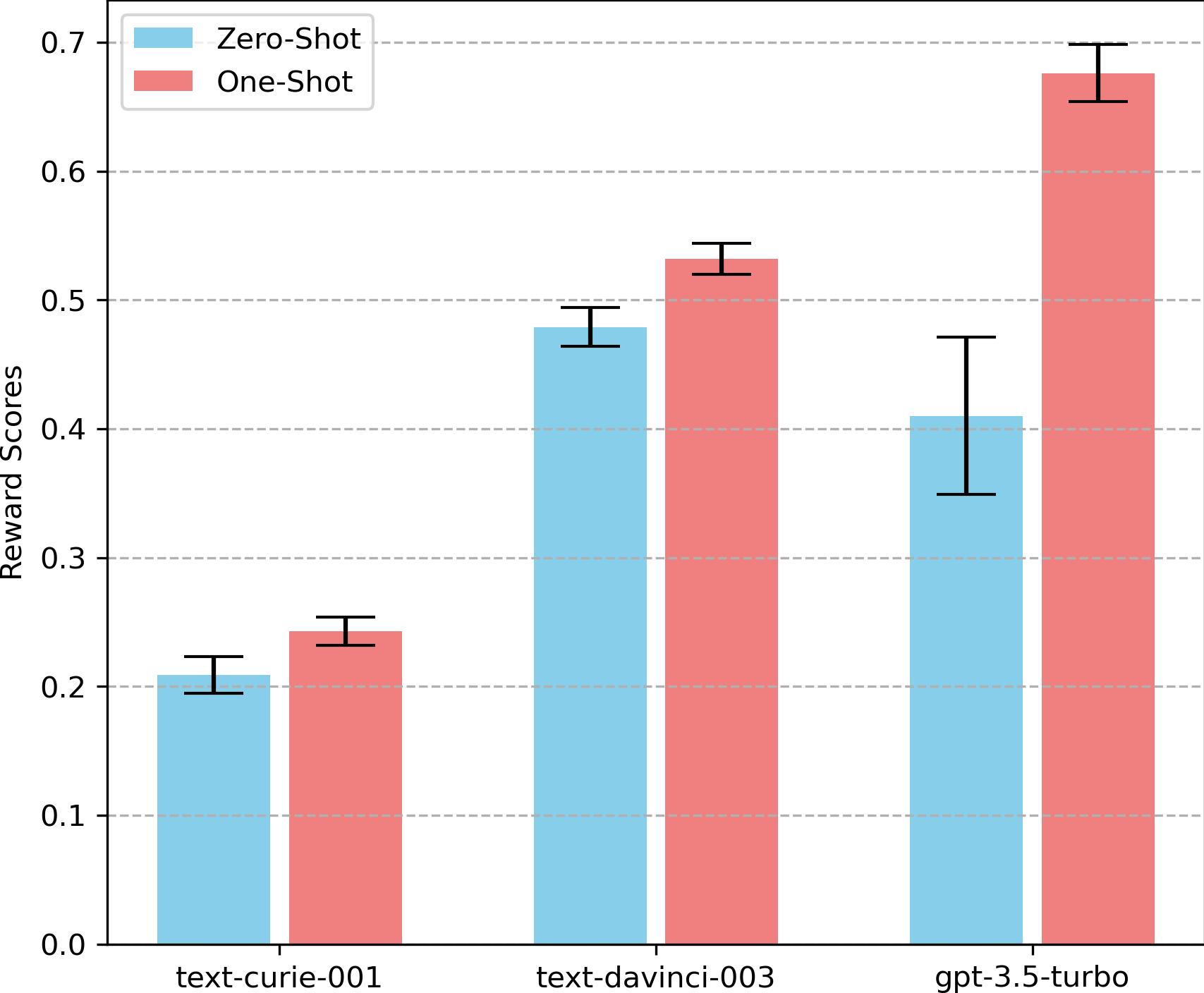}
        \caption{Reward Scores on the Darts-Medium Dataset.}
    \end{subfigure}
    \hfill
    \begin{subfigure}{0.245\textwidth}
        \centering
        \includegraphics[width=\textwidth]{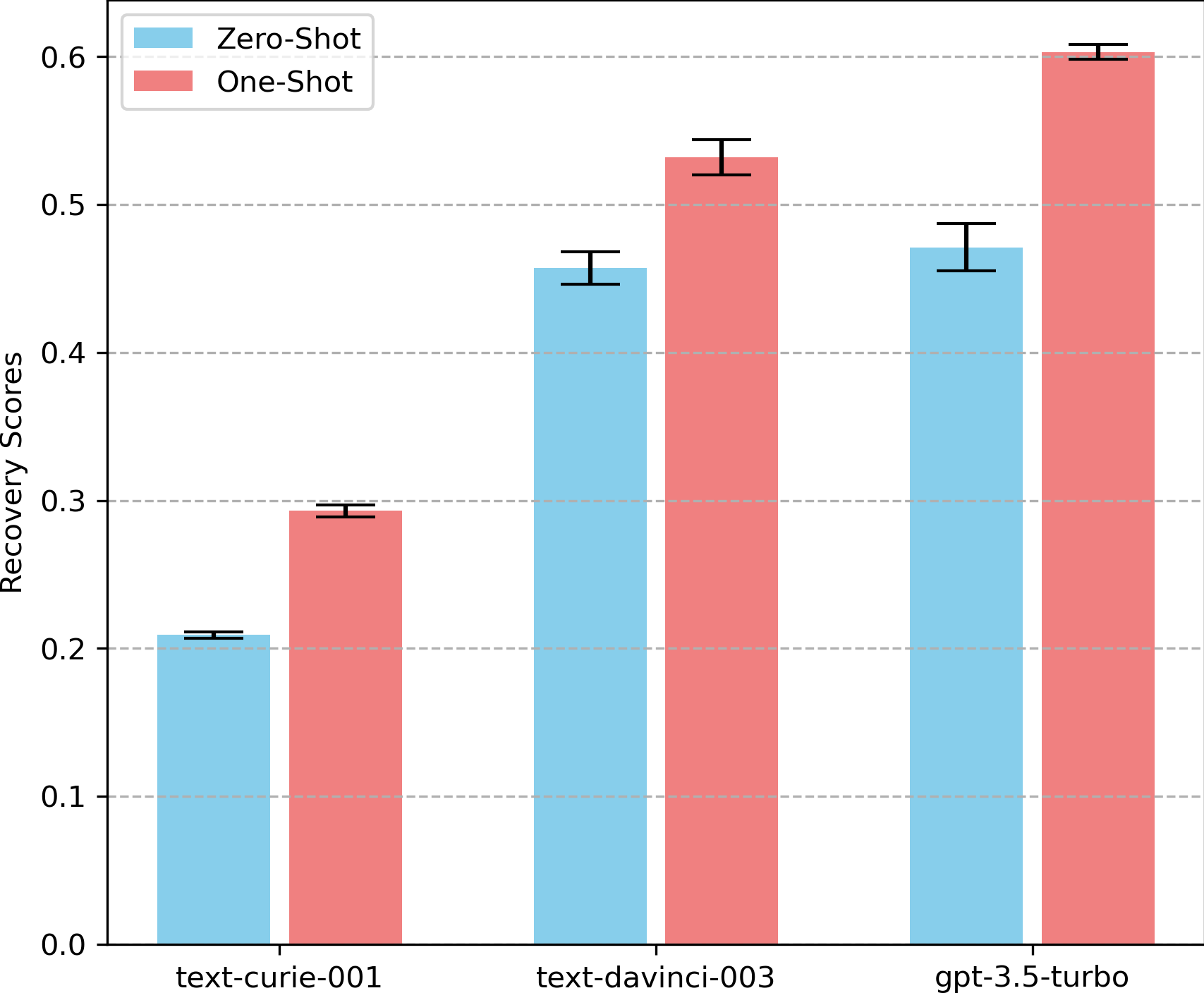}
        \caption{Recovery Scores on the Darts-Medium Dataset.}
    \end{subfigure}
    \hfill
    \begin{subfigure}{0.245\textwidth}
        \centering
        \includegraphics[width=\textwidth]{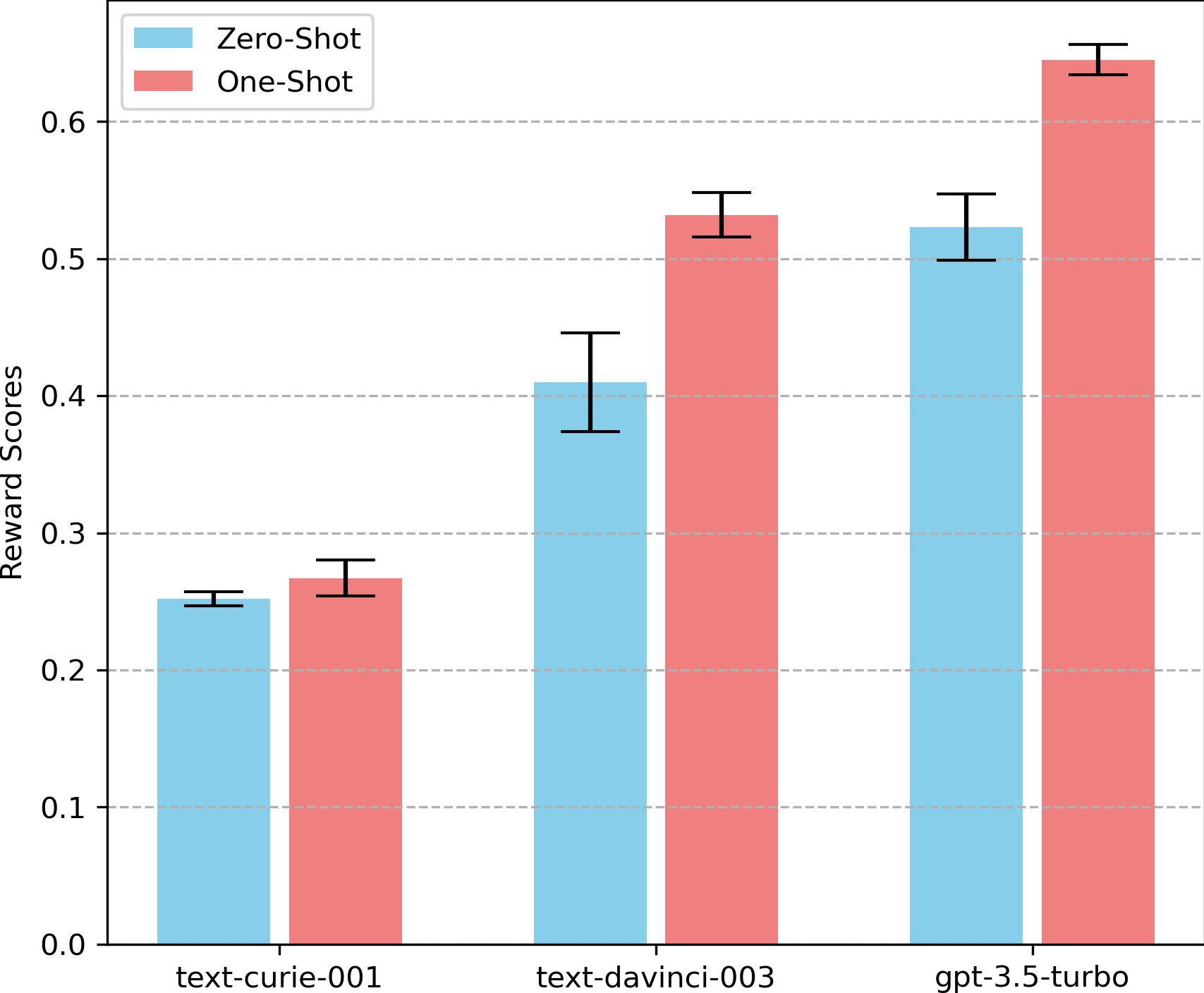}
        \caption{Reward Scores on the PubMed Dataset.}
    \end{subfigure}
    \hfill
    \begin{subfigure}{0.245\textwidth}
        \centering
        \includegraphics[width=\textwidth]{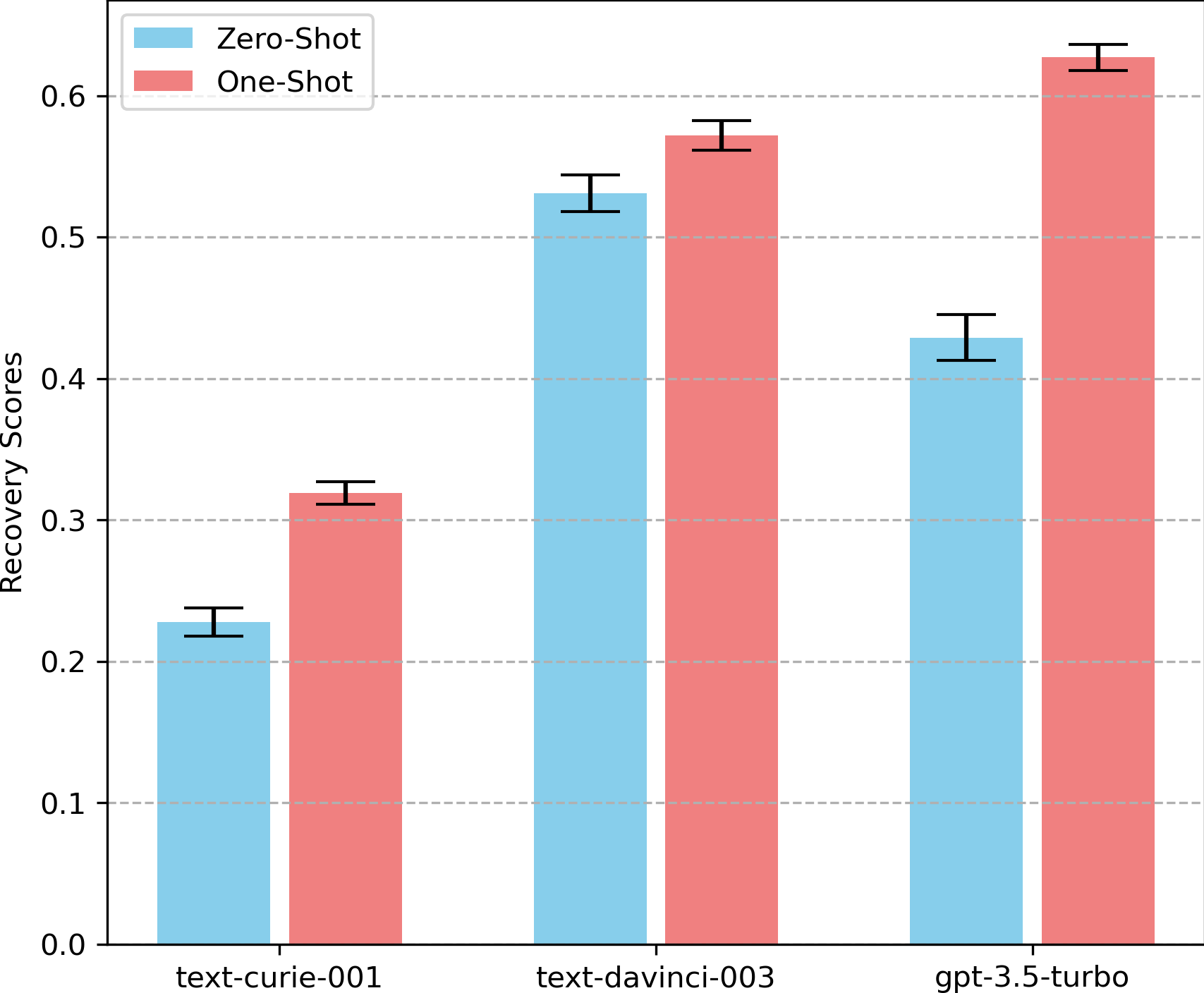}
        \caption{Recovery Scores on the PubMed Dataset.}
    \end{subfigure}
    \caption{The role of the one-shot template navigation: a comparative analysis of one-shot vs. zero-shot approaches. The red bars demonstrate initial the prompt with adding a one-shot template, and the blue bars represent the scores conditinal on the same instruction without a one-shot template. We use gpt-3.5-turbo to conduct the ablation study on both Darts and PubMed datasets.}
    \label{fig:q1}
    \vspace{-0.2in}
\end{figure}
\subsection{Ablation Study}
In the ablation study, we primarily aim to address four key questions for the proposed annotation approach:

\textbf{Q1.} What role does the one-shot template play, and can the same effects be achieved via zero-shot method — generating summaries only with designed instruction?
\textbf{Q2.} If the template work in the self-supervised annotation, what impact do various similarity measurement methods have on the outcomes of one-shot tuning? 
\textbf{Q3.}  How does the initialization of the template affect the results of the optimal one in the one-shot tuning phase?
\textbf{Q4.} Are the generated summary influenced by the hyperparameters of the GPT model itself?

\textbf{A1. The optimal one-shot template improves the annotation performance compared to the zero-shot generation conditional on the same instruction.} To answer \textbf{Q1}, we employ both zero-shot and one-shot approaches to investigate various models and two distinct domain-specific datasets. The evaluation setting follows the steps illustrated in \ref{Evaluation}. Figure \ref{fig:q1} demonstrates a comprehensive comparison of the outcomes of these two generation methods, encompassing the quality of direct generation and the precision of data recovery. We observe that the one-shot mechanism (red), which incorporates template iteration, consistently surpasses the performance of zero-shot (blue) in terms of annotating data quality, regardless of whether we evaluate the reward model or the similarity of the recovered data. Moreover, we observe the standard deviation under different dataset partitions by performing cross-validation during the generation phase. The figure displays smaller red error bars compared to the blue ones, further illustrating that the template enhances the stability and robustness of annotation quality.

\begin{wraptable}{l}{0.5\linewidth} 
  \vspace{-1em}
  \centering
    \centering
    \caption{Impact of different alignment metrics between the Recovered Data and Original Data for the One-Shot Tuning. $\overline{R}$ and $\overline{S}$ are the average value of rewards scores and recovery scores.
    }
    \resizebox{\linewidth}{!}{
    \begin{tabular}{l||cccc||c c}
    \hline
    \textbf{} & STS  & BERT & ROUGE & BLEU & Darts & PubMed\\ \hline
    \multirow{5}{*}{$\overline{R}$} & \checkmark & & & & 52.14 & 58.89 \\ 
                                    &  &   &  \checkmark & & 53.21 & 60.32\\ 
                                    & \checkmark& & \checkmark & & 61.48 & 60.05\\ 
                                    & \checkmark & \checkmark  & \checkmark &  &            62.59   &   65.27   \\
                                    & \checkmark & \checkmark  & \checkmark & \checkmark &            61.17   &   65.96   \\ \hline
    \multirow{5}{*}{$\overline{S}$} & \checkmark &  &  & & 21.18 &  19.04\\ 
                                    & & & \checkmark & & 26.19 &  25.20\\  
                                    & \checkmark & & \checkmark & & 29.44 & 39.55\\ 
                                    & \checkmark & \checkmark  & \checkmark &  &            45.59  &   40.04   \\
                                    & \checkmark & \checkmark  & \checkmark & \checkmark & 44.05 & 41.38\\ \hline
    \end{tabular}
    }
    \label{tab:q2}
    \vspace{-0.1in}
\end{wraptable}
\textbf{A2. Appropriately increasing the measurement metrics benefits the annotation quality.} The feedback process of our proposed iterative algorithm is based on the measure of similarity between the recovered data and the original data, as to the choice of similarity metrics is essential. Considering that both the structured original data and the generated data are sequence information, we apply the most widely used schemes for measuring sequence similarity, ranging from sentence-level measurement methods (BLEU, ROUGE), to embedding-level measurement techniques (STS, BERT). The experimental pipeline for this ablation study adheres to the previous process. 
Table \ref{tab:q2} provides the records of the testing scores by setting different similarity alignments as the feedback value. In the first two rows of the Table \ref{tab:q2},  Initially, we tested the effects of two different types of single similarity calculation functions on the annotated results. The first two rows of each group in the Table \ref{tab:q2} demonstrate this situation. Experimental records suggest that sentence structure has a slightly more positive impact on the results. Moreover, by comparing the last two rows (mixed metric functions) with the first two rows (single metric function) in each group, we observe that whether we directly measure with the reward function or indirectly assess the summary's recovery ability, enhancing the diversity of metric functions can effectively improve the quality of the annotated data. 

\begin{table}[!tbp] 
\caption{Impact of the initial template. From the view of the initial complexity of the data and the initial summary quality. }
\begin{center}
\resizebox{0.8\textwidth}{!}{
\begin{tabular}{l||cc|cc|cc}
\hline
\multirow{2}{*}{\begin{tabular}[c]{@{}l@{}}Initial\\ Template\end{tabular}} & \multicolumn{2}{c|}{Data of 3 operators}                   & \multicolumn{2}{c|}{Data of 5 operators}                   & \multicolumn{2}{c}{Data of 7 operators}                   \\
                                                                            & \multicolumn{1}{l}{Sum. High} & \multicolumn{1}{l|}{Sum. Low} & \multicolumn{1}{l}{Sum. High} & \multicolumn{1}{l|}{Sum. Low} & \multicolumn{1}{l}{Sum. High} & \multicolumn{1}{l}{Sum. Low} \\ \hline
Iteration                                                                      & 4                           & 8                            & 5                           & 8                            & 4                           & 7                           \\
Similarity                                                                        & 0.653                       & 0.652                        & 0.641                       & 0.655                        & 0.652                       & 0.659                       \\ \hline
\end{tabular}
}
\label{tab:q3}
\end{center}
\vspace{-2em}
\end{table}

\textbf{A3. High-quality initial templates lead to quicker convergence, but the similarity scores of the last iteration exhibit low variance.} During the initial phase of one-shot tuning, we made observations in the running records which indicated that the length and quality of the initialized templates had an impact on the training process. To investigate this further, we designed three templates of the Darts dataset with varying levels of complexity:  Darts dataset generated by 3 operators,  5 operators, and 7 operators. Additionally, we artificially marked 2 types of summary qualities: Sim. High and Sim. Low. This resulted in a total of 6 cases. To ensure consistent experimental conditions, each template is applied 50 times during the one-shot stage. Moreover, the temperature of GPT was set to 0 to maintain the stability of summary generation in each round. Table \ref{tab:q3} demonstrates the influence of six cases of various settings on the initialization. The polyline of the warm colour group represents the iterative process of high-quality summaries. We can observe from this that its convergence time is relatively early compared to the initial template that with lower-quality summaries. The score at the algorithm termination clearly shows that the final generated optimal template possesses a close similarity score, which indicates that initialization mainly affects the iteration number but does not apply to the performance of the optimal template. 

\begin{figure}[!tbp]
    \centering
    \begin{subfigure}{0.245\textwidth}
        \centering
        \includegraphics[width=\textwidth]{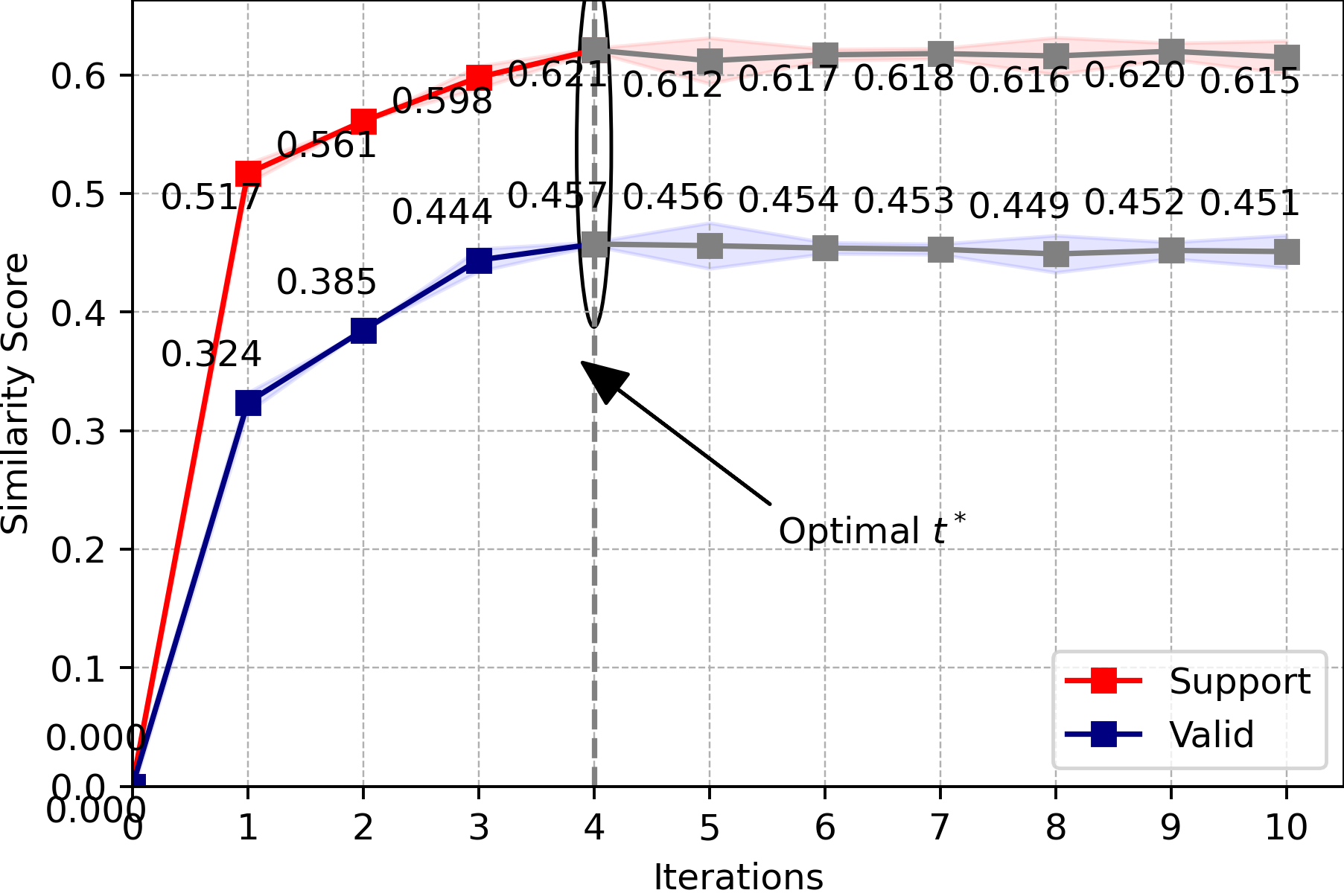}
        \caption{T=0 of text-curie-001 on Darts-Medium;}
    \end{subfigure}
    \hfill
    \begin{subfigure}{0.245\textwidth}
        \centering
        \includegraphics[width=\textwidth]{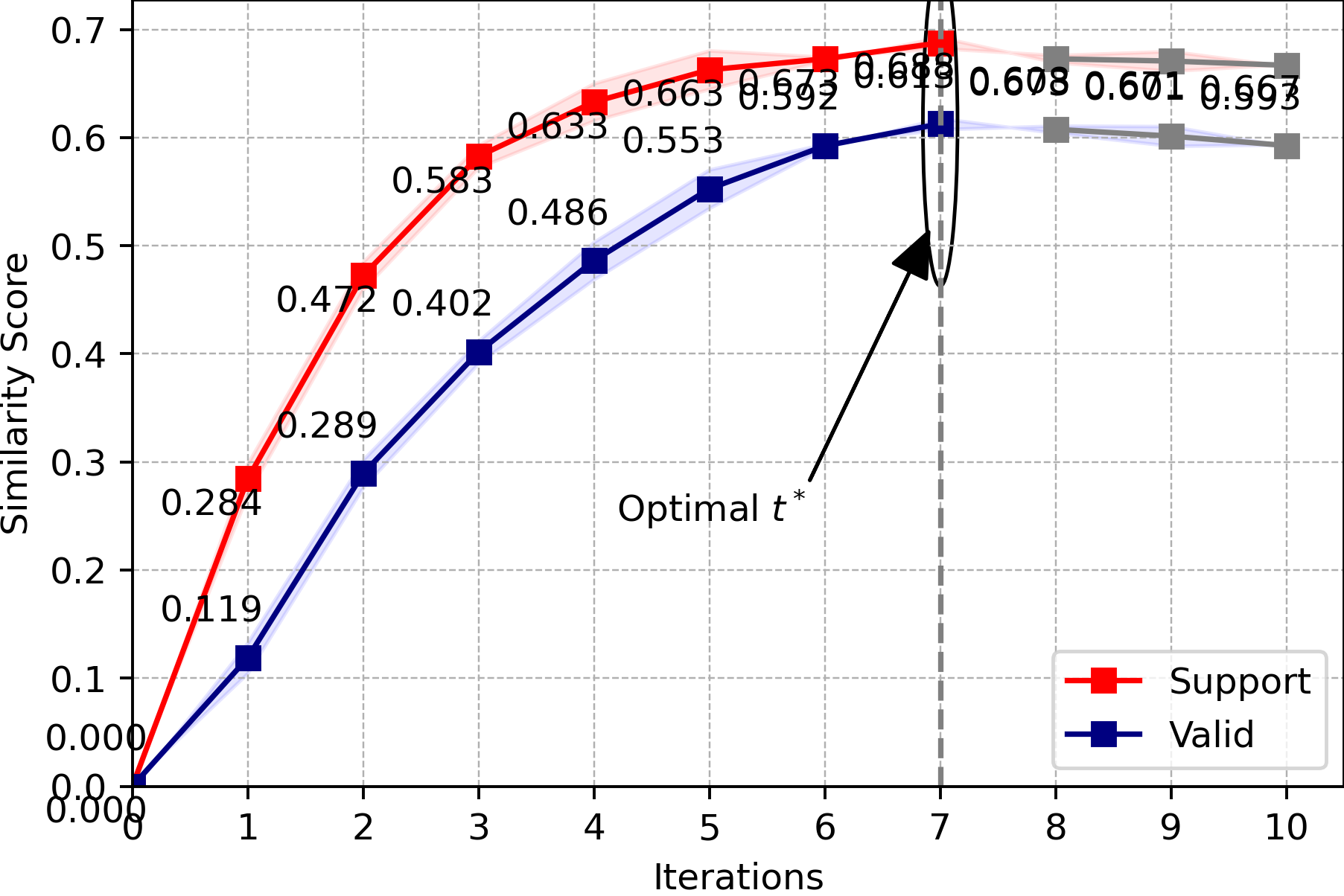}
        \caption{T=1 of text-curie-001 on Darts-Medium;}
    \end{subfigure}
    \hfill
    \begin{subfigure}{0.245\textwidth}
        \centering
        \includegraphics[width=\textwidth]{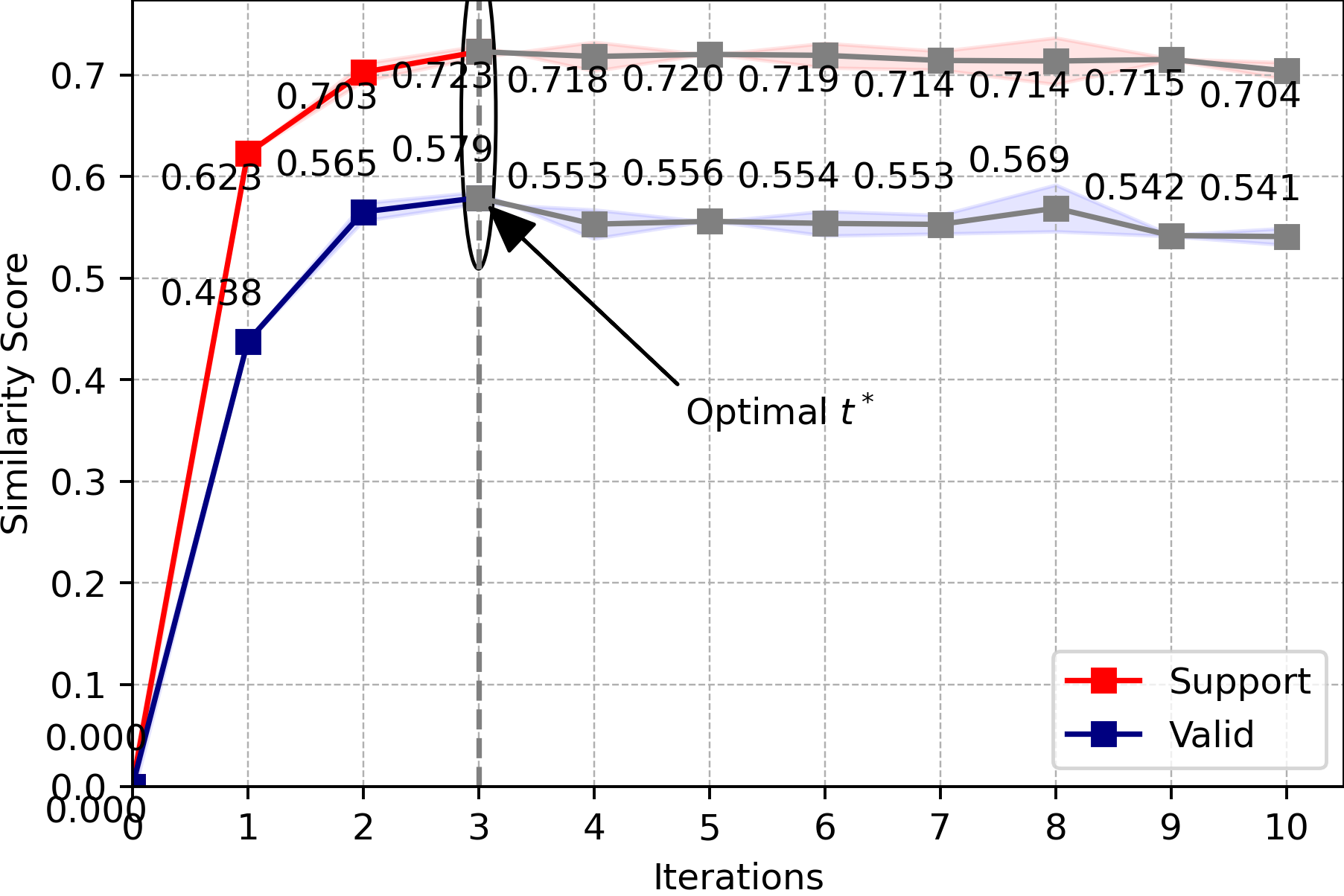}
        \caption{T=0 of gpt-3.5-turbo on the Darts-Medium;}
    \end{subfigure}
    \hfill
    \begin{subfigure}{0.245\textwidth}
        \centering
        \includegraphics[width=\textwidth]{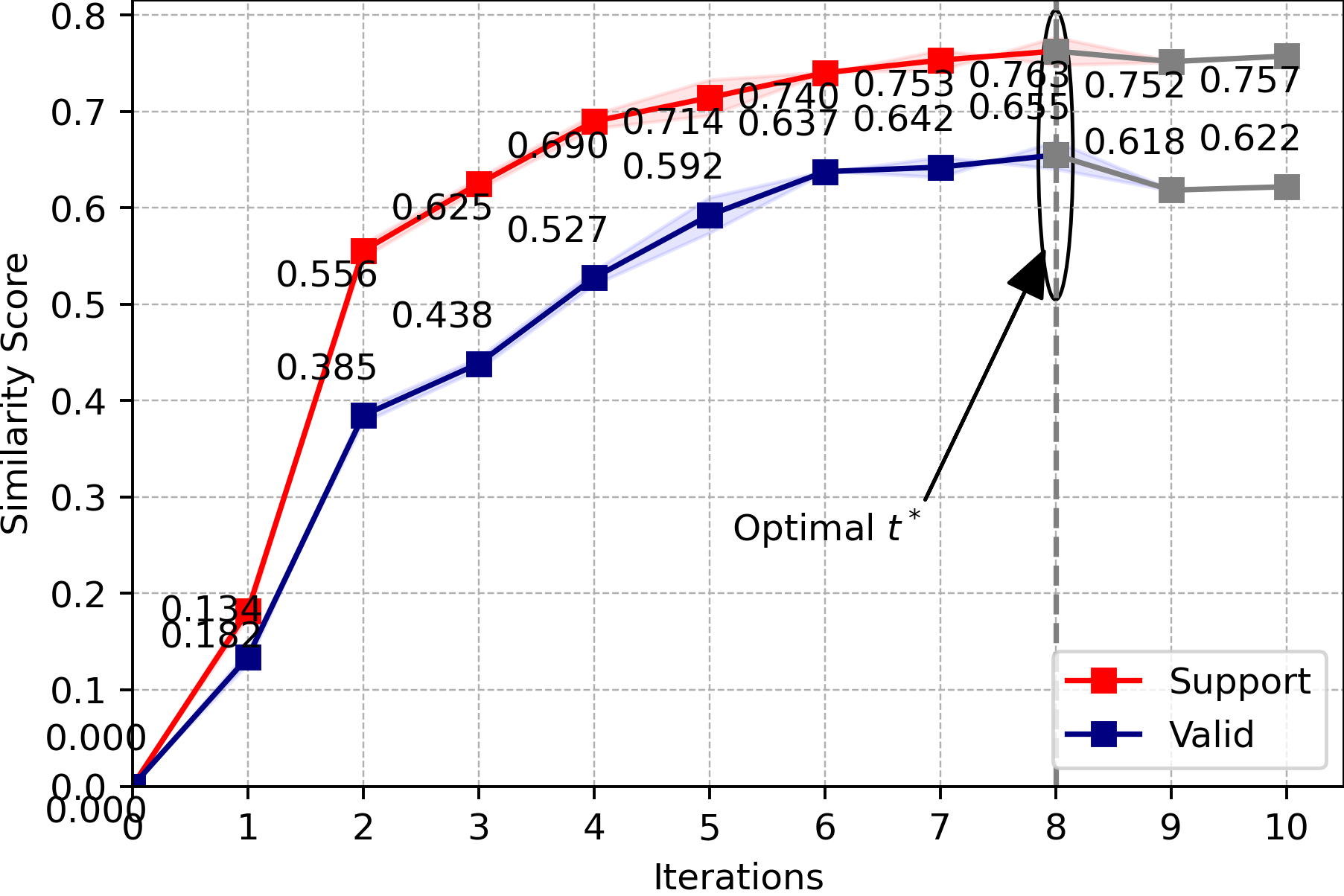}
        \caption{T=1 of gpt-3.5-turbo on the Darts-Medium;}
    \end{subfigure}
        \begin{subfigure}{0.245\textwidth}
        \centering
        \includegraphics[width=\textwidth]{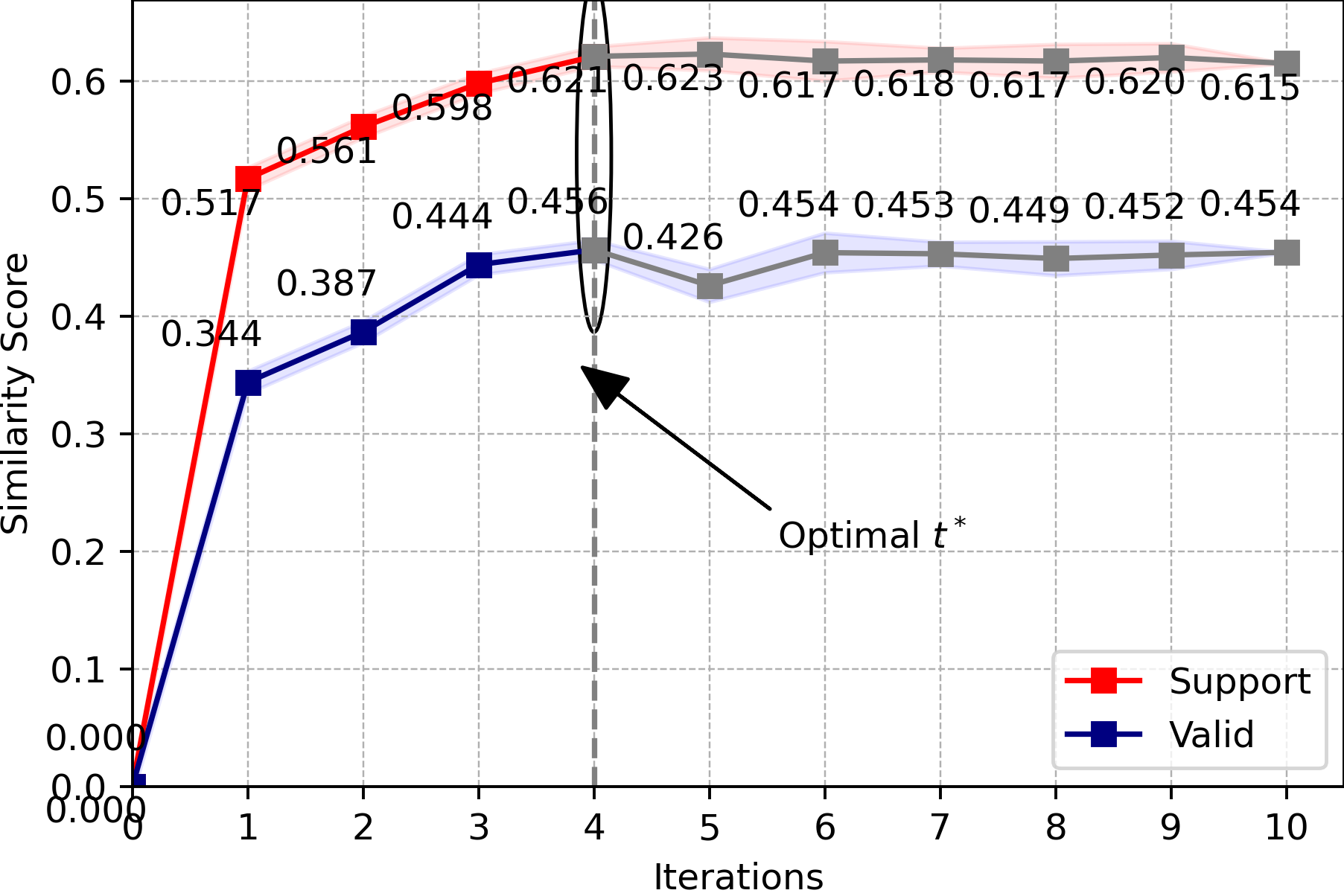}
        \caption{T=0 of davinci on the Darts-Medium;}
    \end{subfigure}
    \hfill
    \begin{subfigure}{0.245\textwidth}
        \centering
        \includegraphics[width=\textwidth]{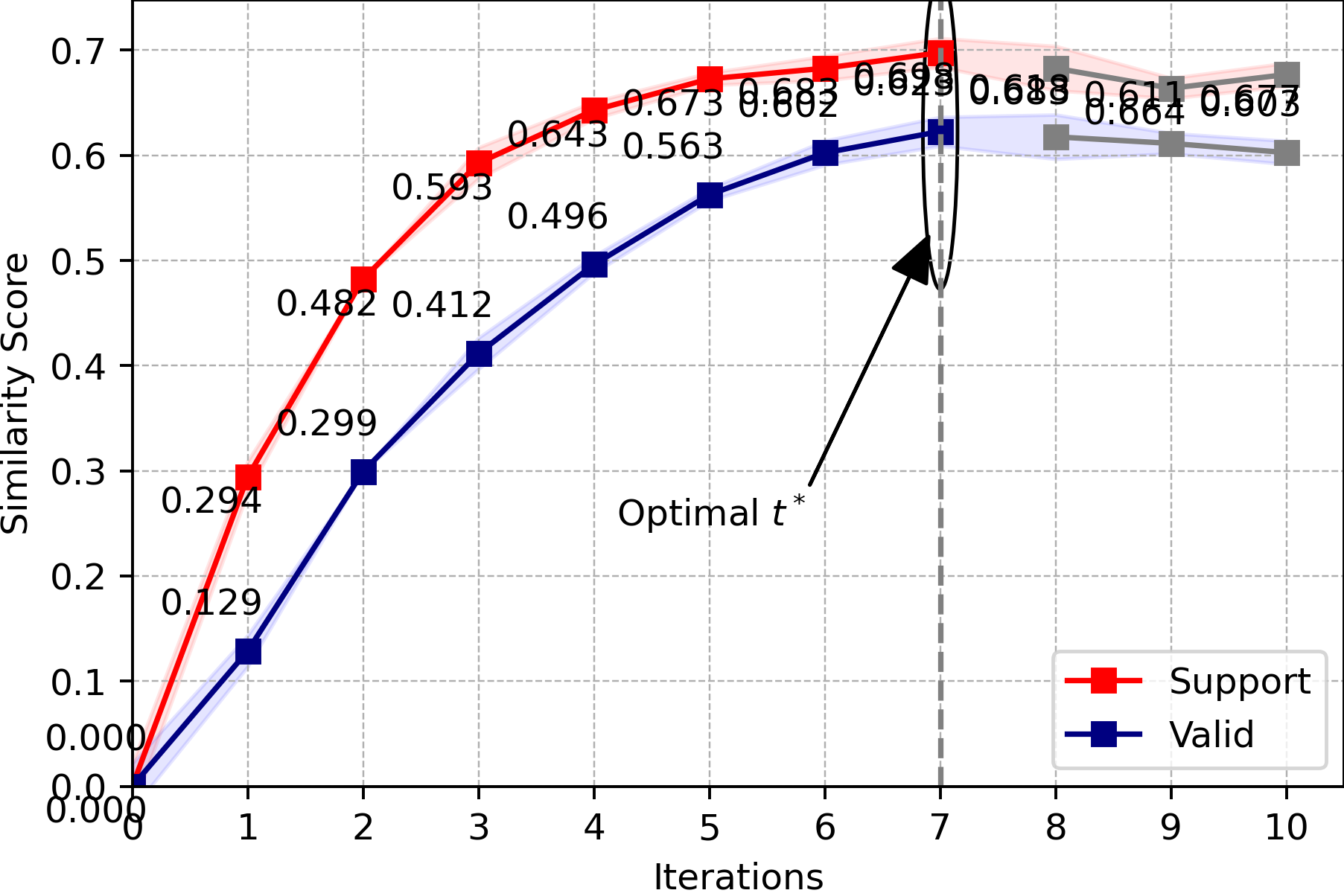}
        \caption{T=1 of davinci on the Darts-Medium;}
    \end{subfigure}
    \hfill
    \begin{subfigure}{0.245\textwidth}
        \centering
        \includegraphics[width=\textwidth]{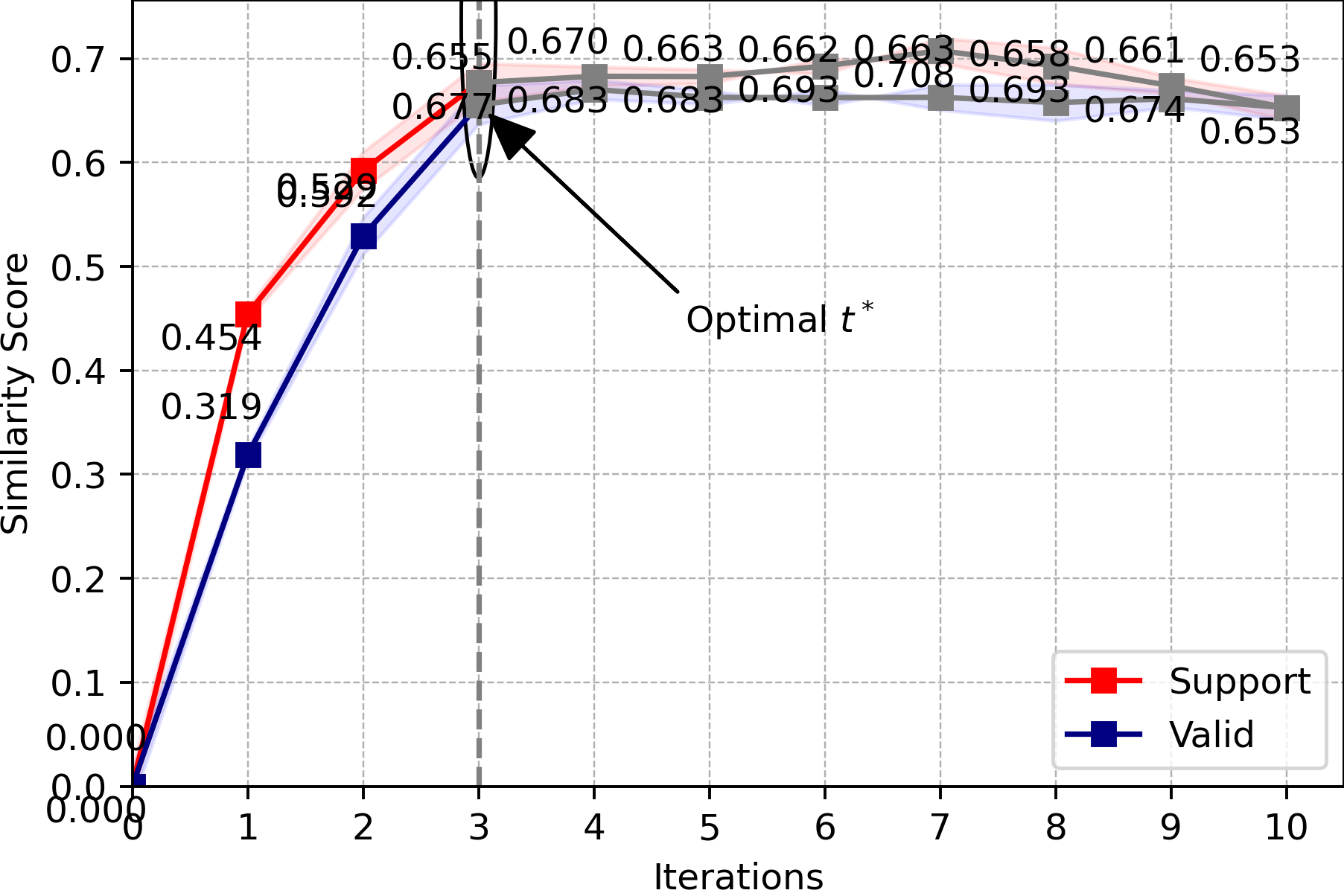}
        \caption{T=0 of text-davinci-003 on the Darts-Medium;}
    \end{subfigure}
    \hfill
    \begin{subfigure}{0.245\textwidth}
        \centering
        \includegraphics[width=\textwidth]{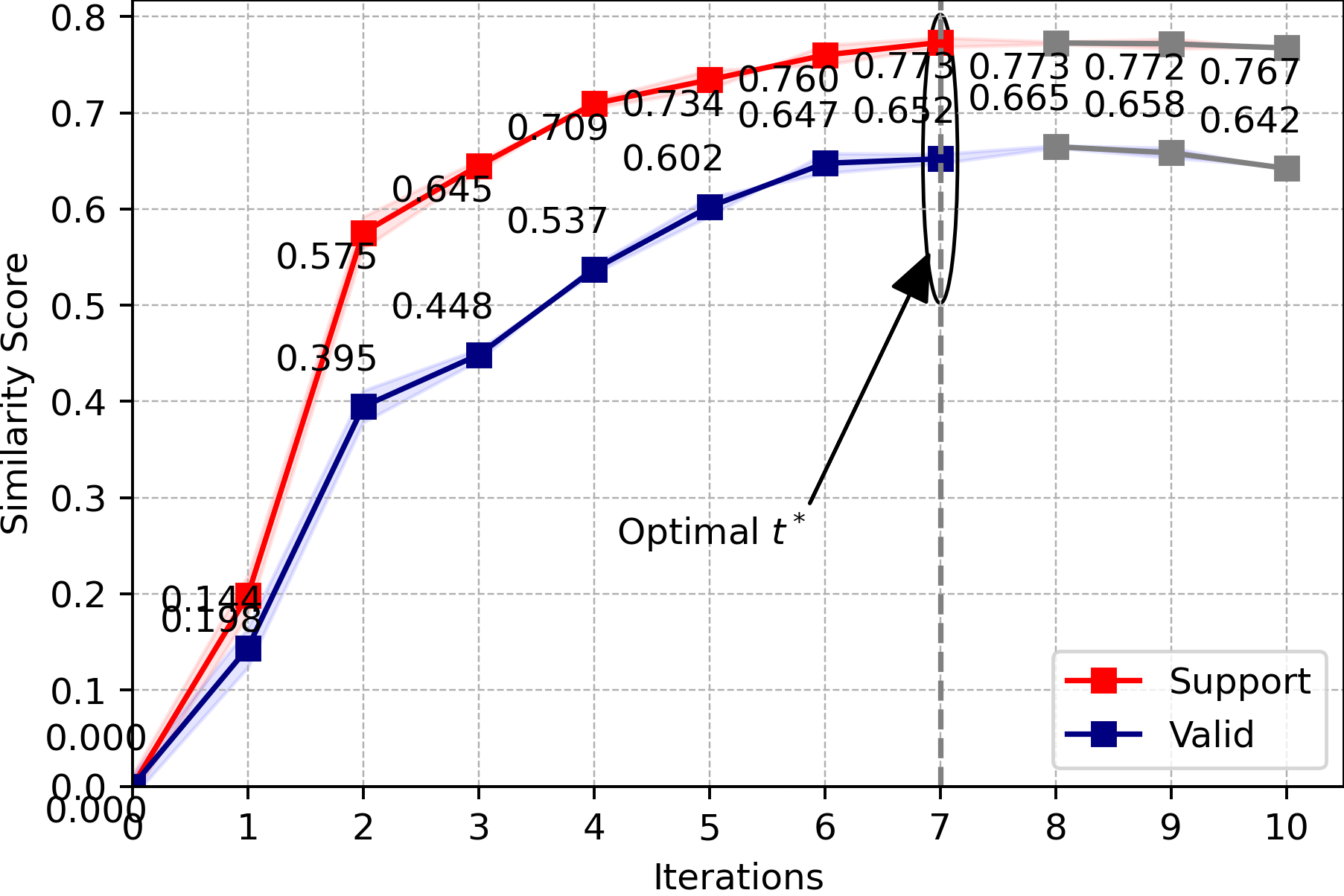}
        \caption{T=1 of text-davinci-003 on the Darts-Medium;}
    \end{subfigure}
        \begin{subfigure}{0.245\textwidth}
        \centering
        \includegraphics[width=\textwidth]{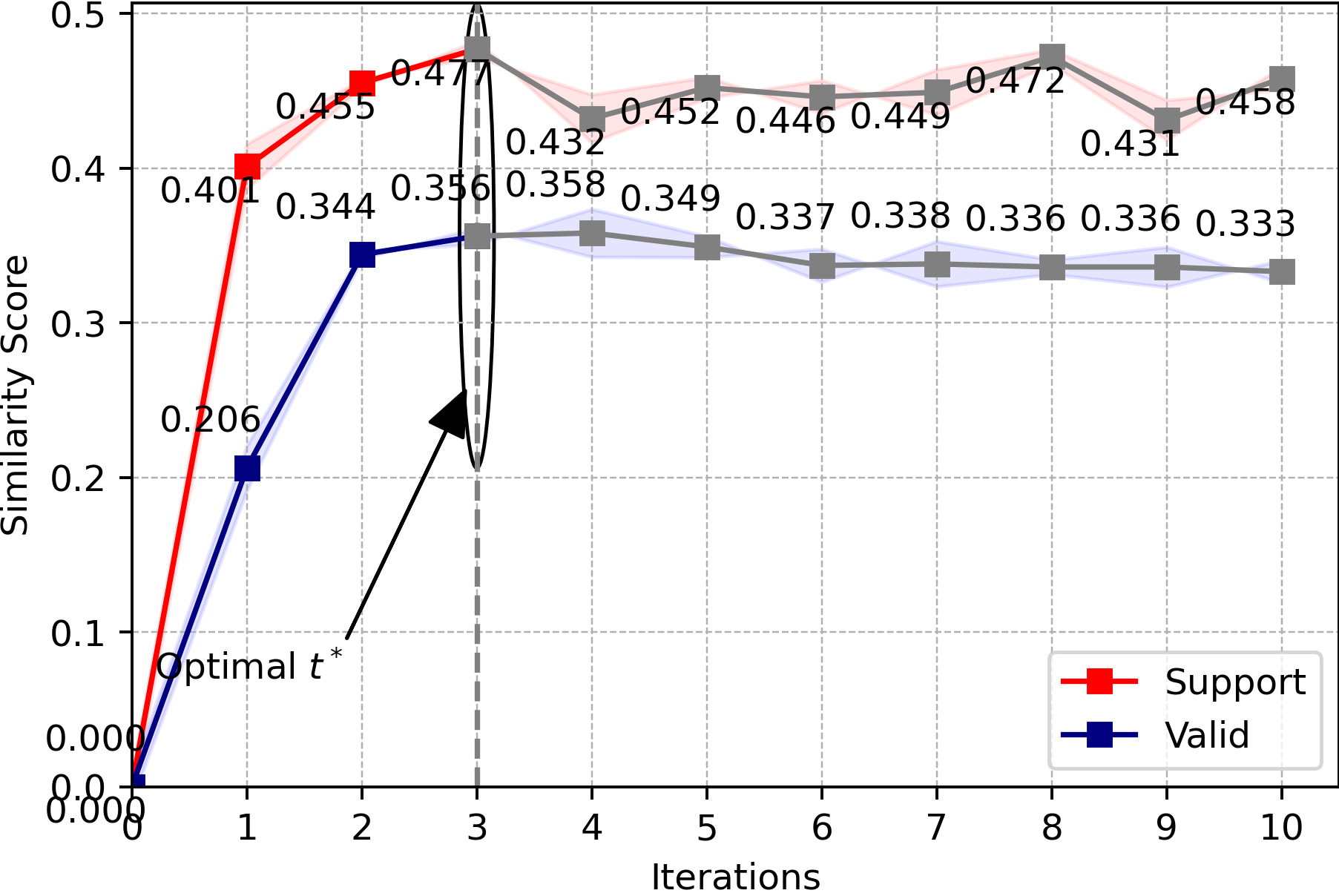}
        \caption{T=0 of text-curie-001 on the PubMed;}
    \end{subfigure}
    \hfill
    \begin{subfigure}{0.245\textwidth}
        \centering
        \includegraphics[width=\textwidth]{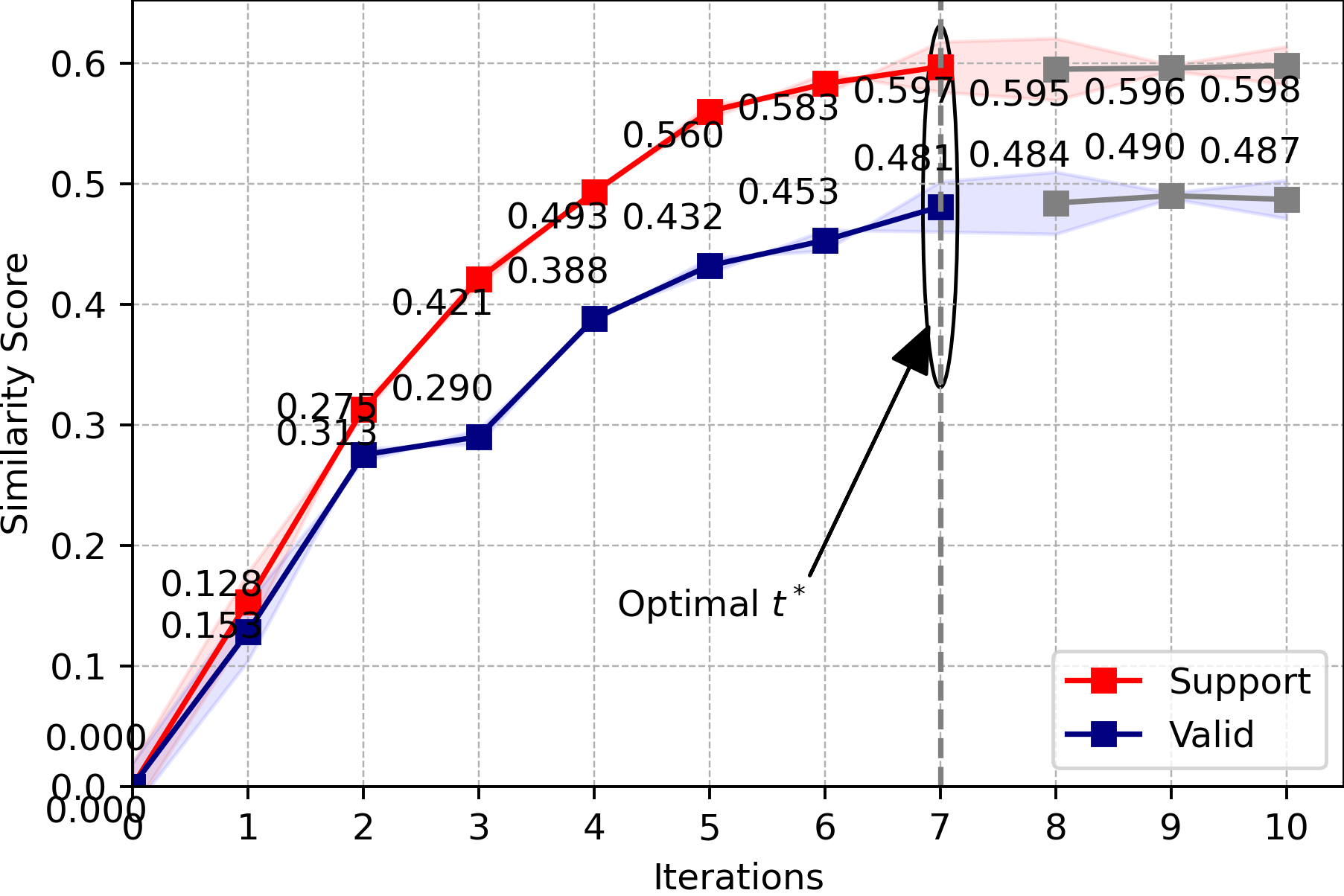}
        \caption{T=1 of text-curie-001 on the PubMed;}
    \end{subfigure}
    \hfill
    \begin{subfigure}{0.245\textwidth}
        \centering
        \includegraphics[width=\textwidth]{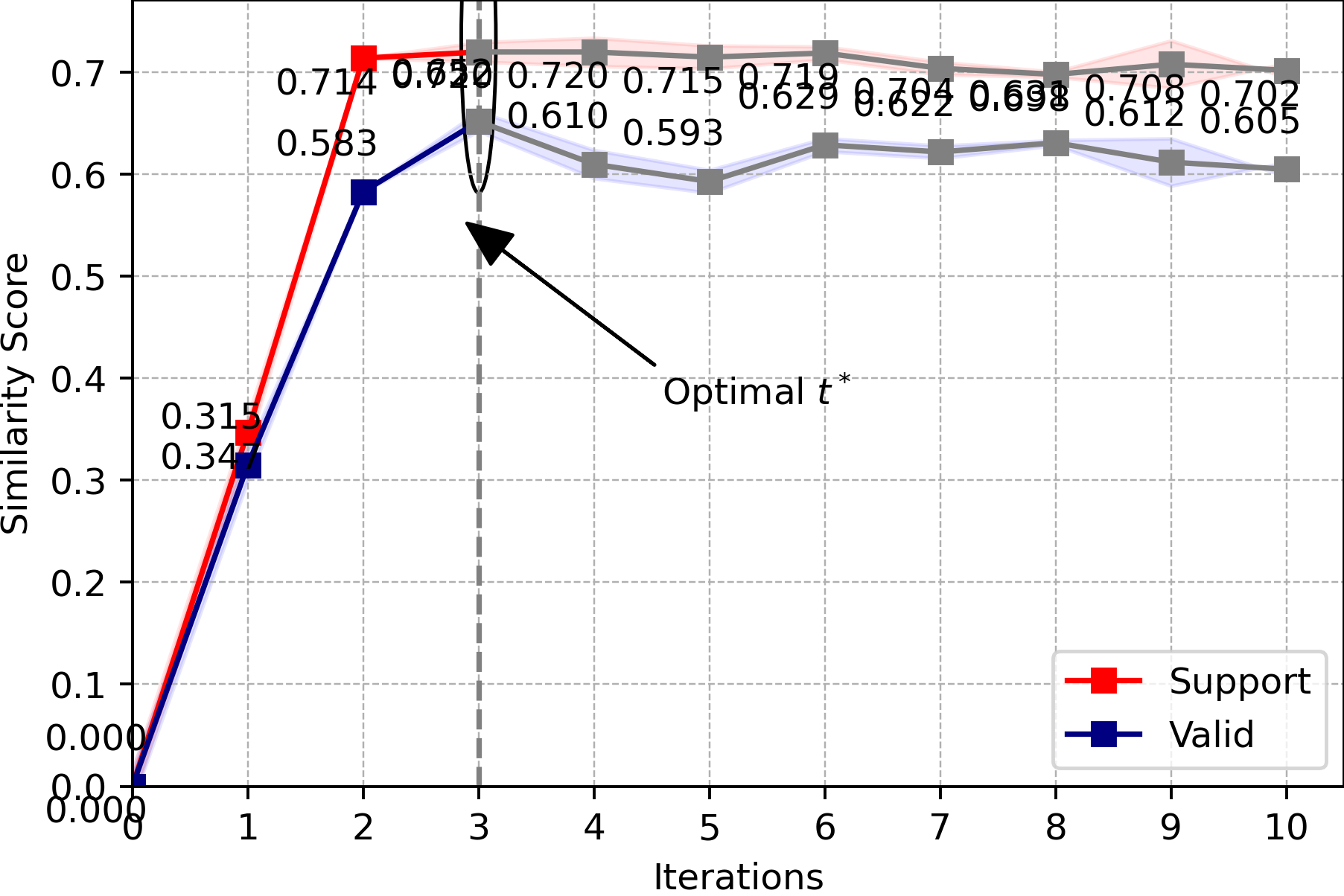}
        \caption{T=0 of gpt-3.5-turbo on the PubMed;}
    \end{subfigure}
    \hfill
    \begin{subfigure}{0.245\textwidth}
        \centering
        \includegraphics[width=\textwidth]{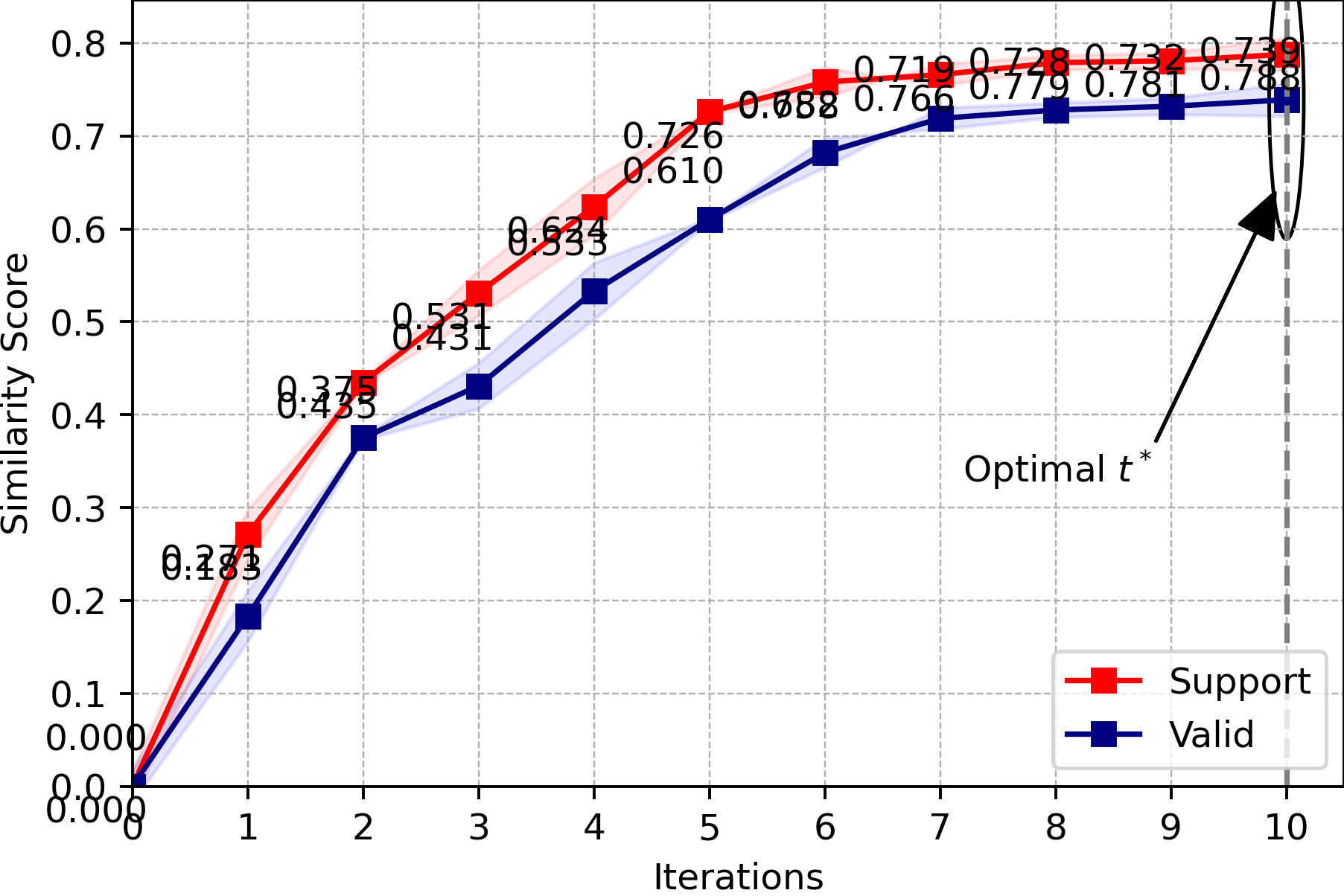}
        \caption{T=1 of gpt-3.5-turbo on the PubMed;}
    \end{subfigure}
        \begin{subfigure}{0.245\textwidth}
        \centering
        \includegraphics[width=\textwidth]{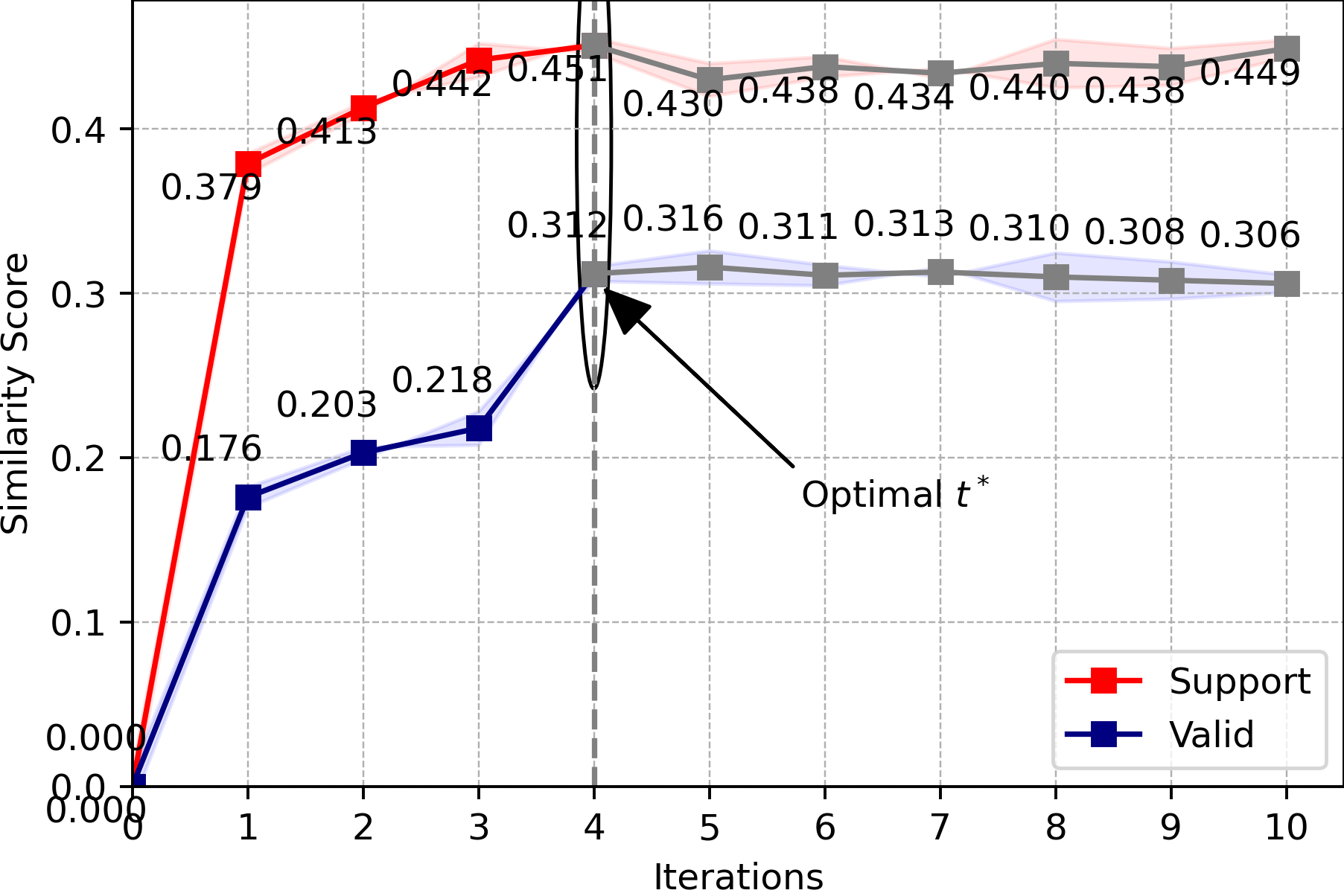}
        \caption{T=0 of davinci on the PubMed;}
    \end{subfigure}
    \hfill
    \begin{subfigure}{0.245\textwidth}
        \centering
        \includegraphics[width=\textwidth]{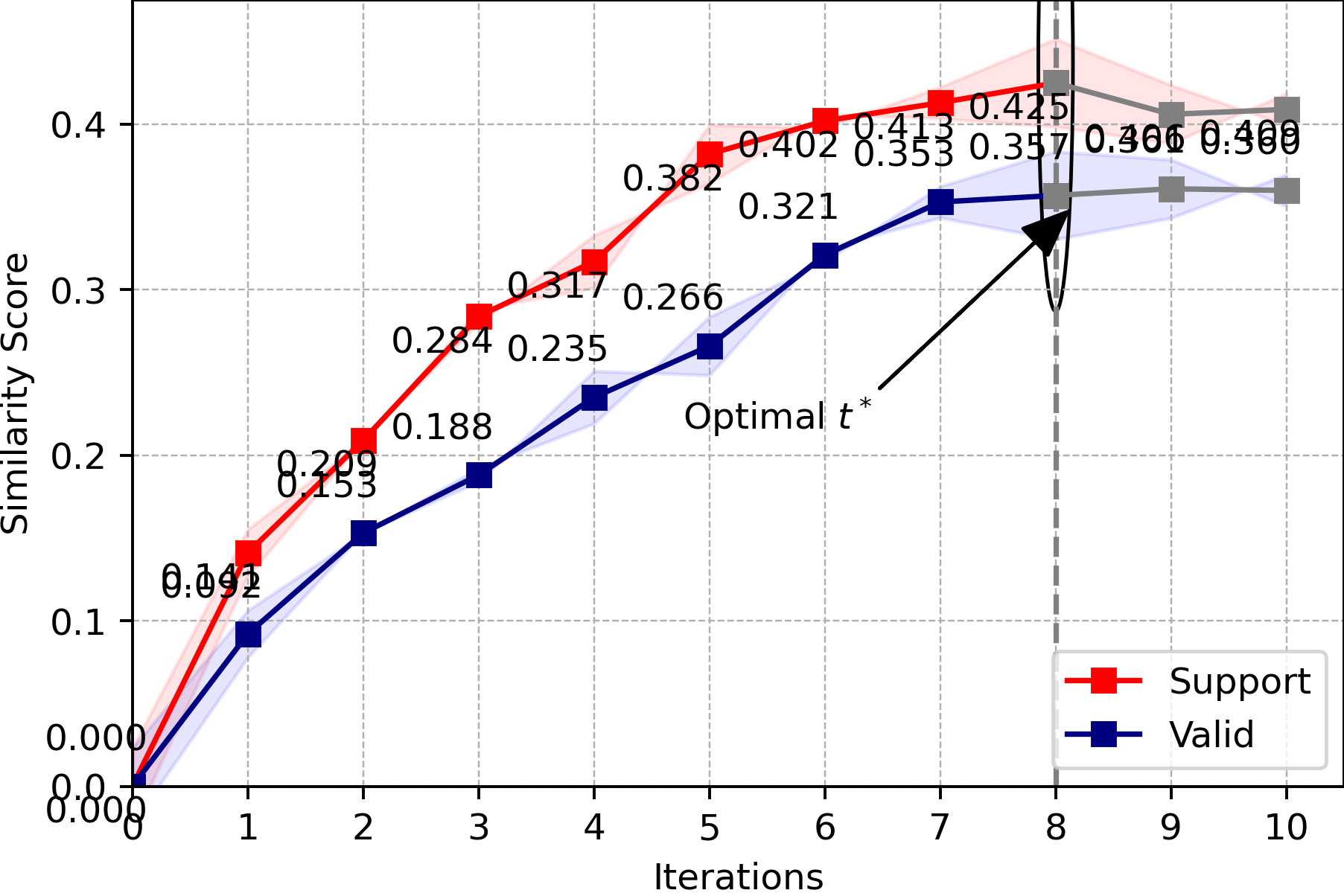}
        \caption{T=1 of davinci on the PubMed;}
    \end{subfigure}
    \hfill
    \begin{subfigure}{0.245\textwidth}
        \centering
        \includegraphics[width=\textwidth]{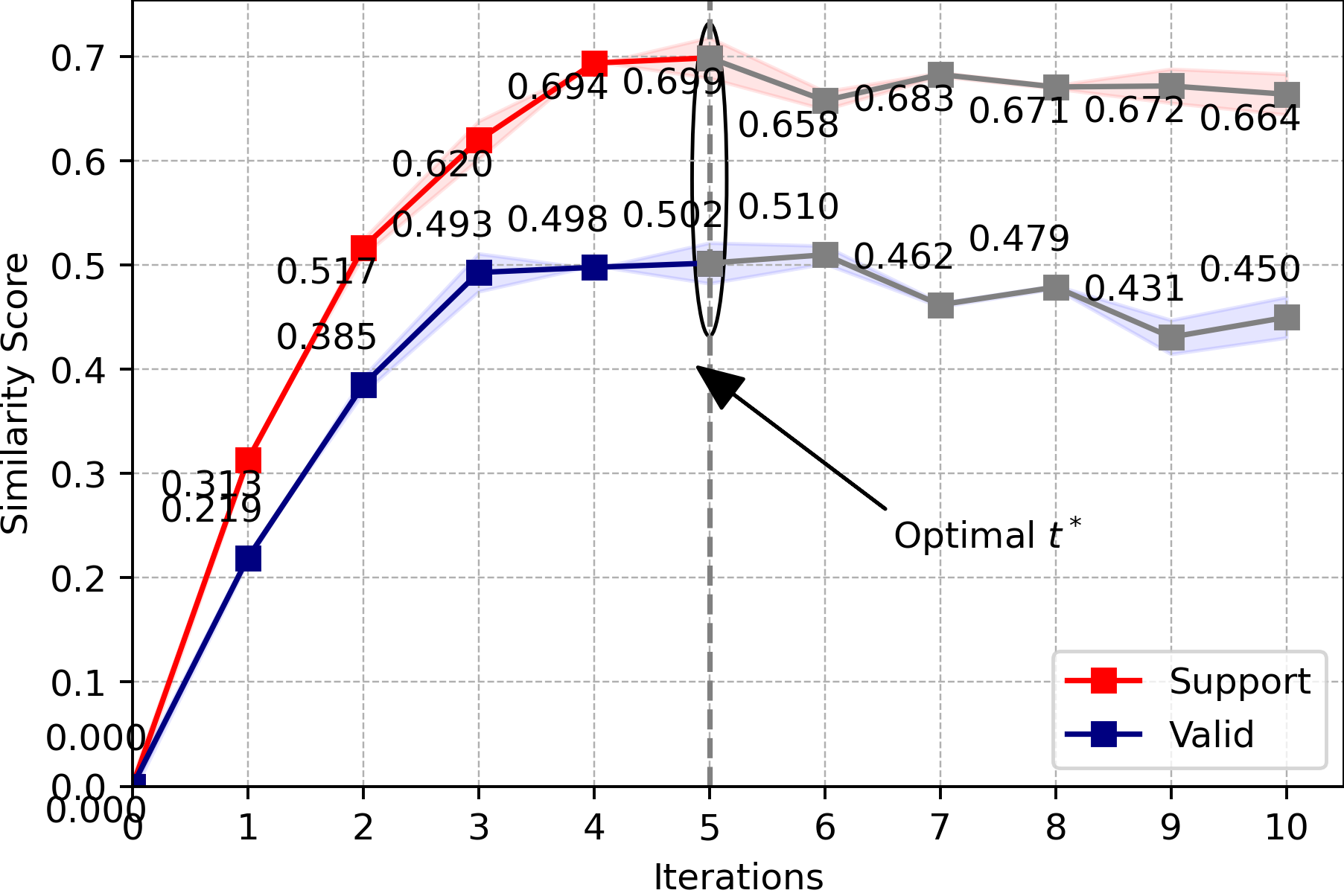}
        \caption{T=0 of text-davinci-003 on the PubMed;}
    \end{subfigure}
    \hfill
    \begin{subfigure}{0.245\textwidth}
        \centering
        \includegraphics[width=\textwidth]{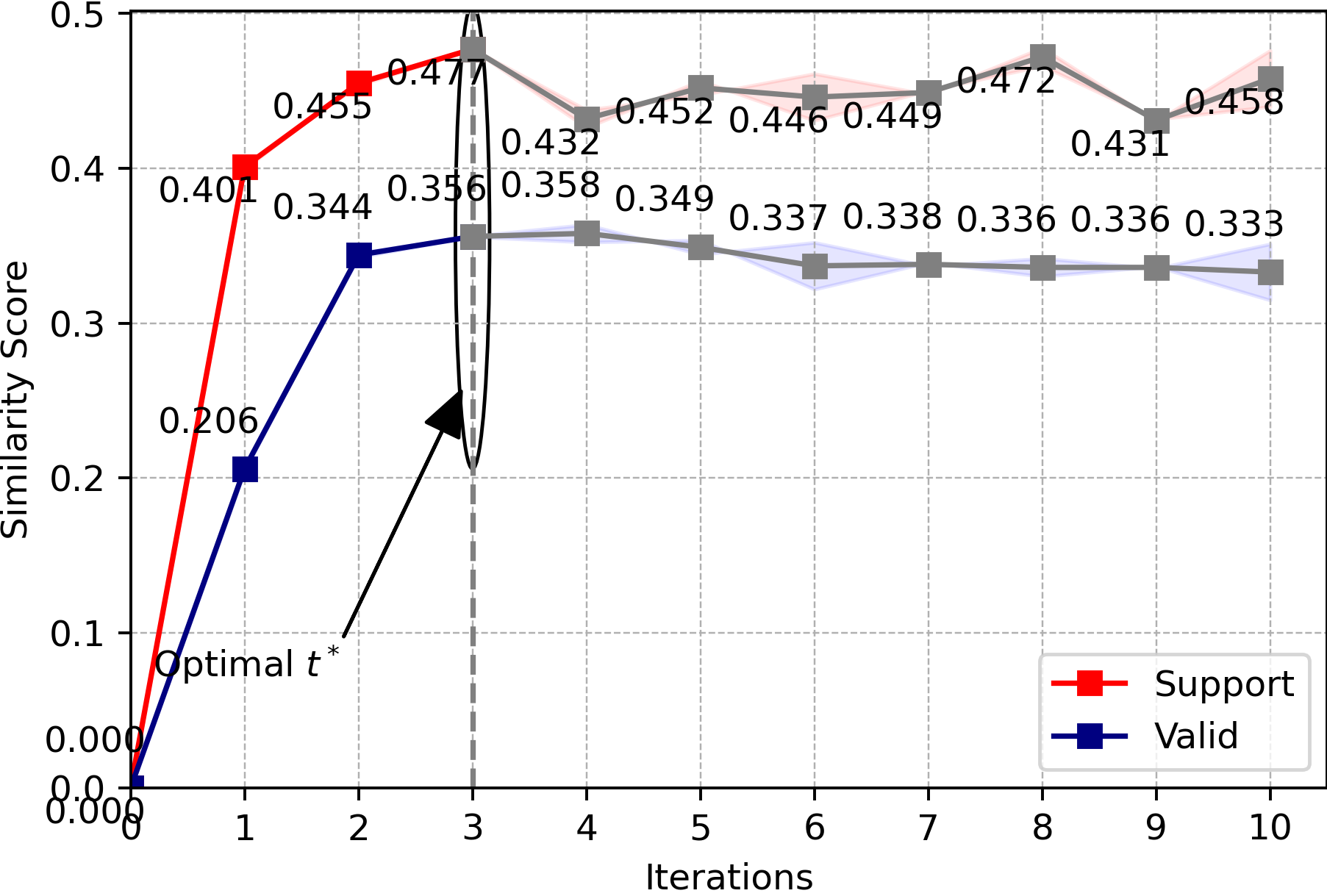}
        \caption{T=1 of text-davinci-003 on the PubMed;}
    \end{subfigure}
    \caption{The Impact of the model's temperature. The red lines represent the iterative records of similarity score by current support data and the blue lines trace the recovery scores on the validation set.
    The arrow points the average number of the iteration to find the optimal template. }
    \label{fig:q4}
    \vspace{-2em}
\end{figure}
\textbf{A4. The temperature of the model influences the iterative search for the optimal template.}
In order to comprehensively investigate the impact of temperature hyperparameters on the performance of different models, we conducted a series of experiments focusing on the convergence properties of four baseline models under varying temperature settings in different environments. To ensure the robustness of our findings, we performed each run 30 times, tracing the standard deviation of the similarity values generated at each instance. Figure \ref{fig:q4} presents the experimental outcomes for the amalgamation of two temperature parameters across the four models. These results reveal a consistent trend across all models: when the temperature is set to 0, the iterative search process terminates prematurely at the third or fourth iteration. This suggests an inclination towards a limited exploration space, leading to the generation of less diverse outputs. On the other hand, elevating the temperature value to 1 during the one-shot tuning stage demonstrated an interesting outcome. Although each iteration's pace was reduced, this facilitated a broader array of alternatives for subsequent template generation. This indicates an expansion of the exploration space, allowing for the generation of more diverse and potentially creative solutions. We also observed that the final scores were higher when the temperature was set to 1, as compared to 0. This finding 
indicates that the optimal templates could be obtained by appropriately tuning the temperature hyperparameter.

\section{Case Study}
\begin{figure}[!tbp]
    \centering
    \includegraphics[width=\linewidth]{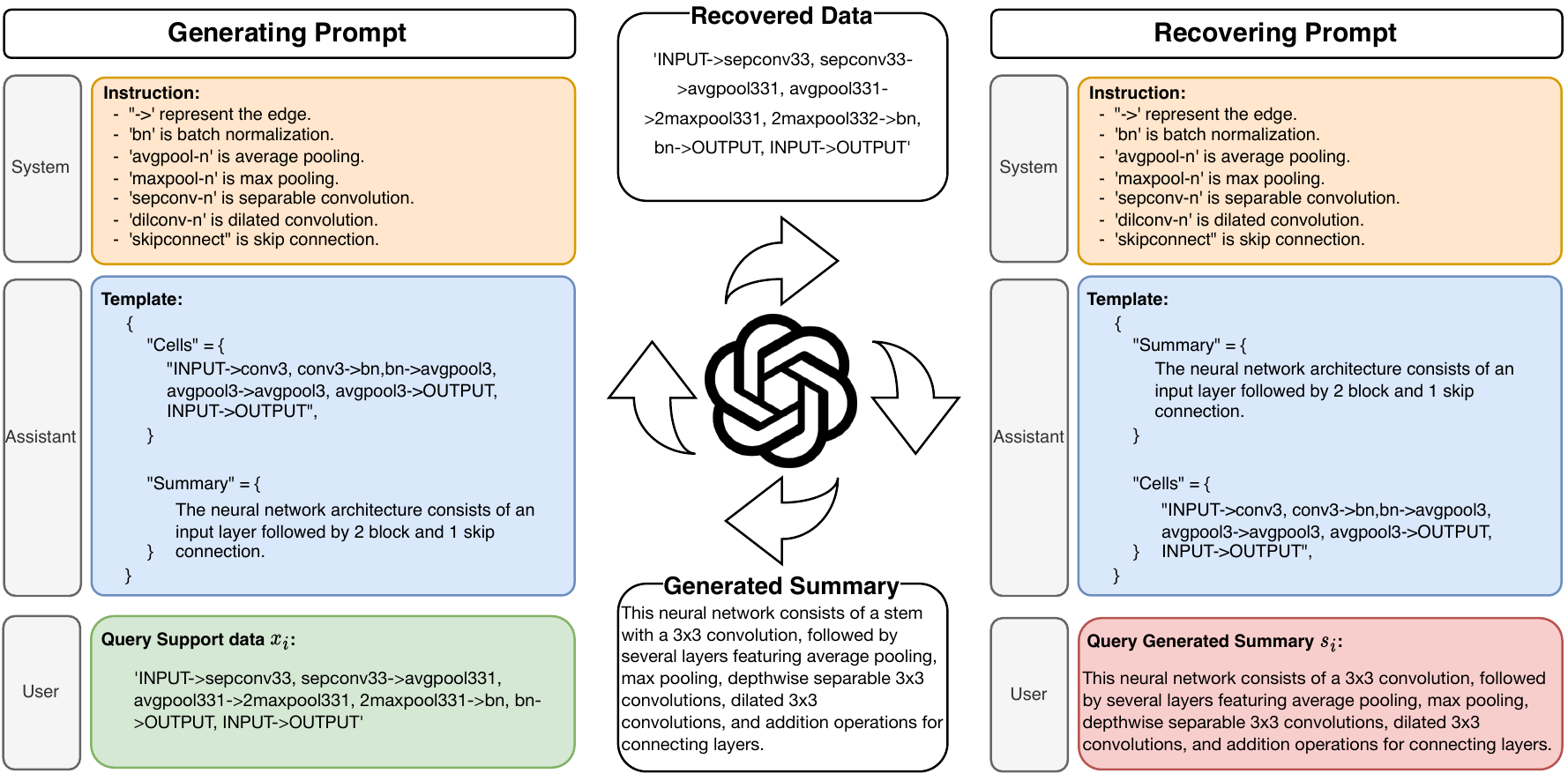}
    \caption{Darts datasets: A case of the generating-recovering annotation.}
  \label{fig:case1}
\end{figure}

Here we provide two cases to elaborate how GPT self-supervised annotating the complex structured data on the Darts and PubMed dataset. We first demonstrate the prompt for generating summary and recovering data, and then we demonstrate the pipelines of our approach.
\paragraph{Prompts}
The generating prompt comprises three elements: {encoding instruction, template, query data}, while the recovering prompt contains {decoding instruction, template, query summary}. Detailed explanations of these instructions can be found in Appendix \ref{sec:Appendix}.
\paragraph{Pipeline}
We use the case \ref{fig:case1} on the Darts dataset to elaborate on the pipeline of our annotation approach.
\paragraph{Step 1.} Initialize a template composed of cells and a summary.
\paragraph{Step 2.} Sample a set of cells from the support set.
\paragraph{Step 3.} Concatenate the <encoding instruction, template, cells> into one message.
\paragraph{Step 4.} Send the message to GPT and receive a summary response.
\paragraph{Step 5.} Concatenate the <decoding instruction, template, summary> into one message.
\paragraph{Step 6.} Send the message to GPT and receive a response of recovered cells.
\paragraph{Step 7.} Compute the similarity between the recovered cells and original cells. If it is larger than the previous record, update the score and treat the sampled data and generated summary as a temporary template.
\paragraph{Step 8.} Evaluate the temporary template on the valid set by repeating the above process.
\paragraph{Step 9.} If the average scores on the validation dataset also surpass previous records, update the best valid score and replace the current template with the temporary template.
\paragraph{Step 10.} Iteratively repeat \textbf{Step 2.} to \textbf{Step 9.} until the maximum number of iterations is reached.

\begin{figure}[H]
    \centering
    \includegraphics[width=\linewidth]{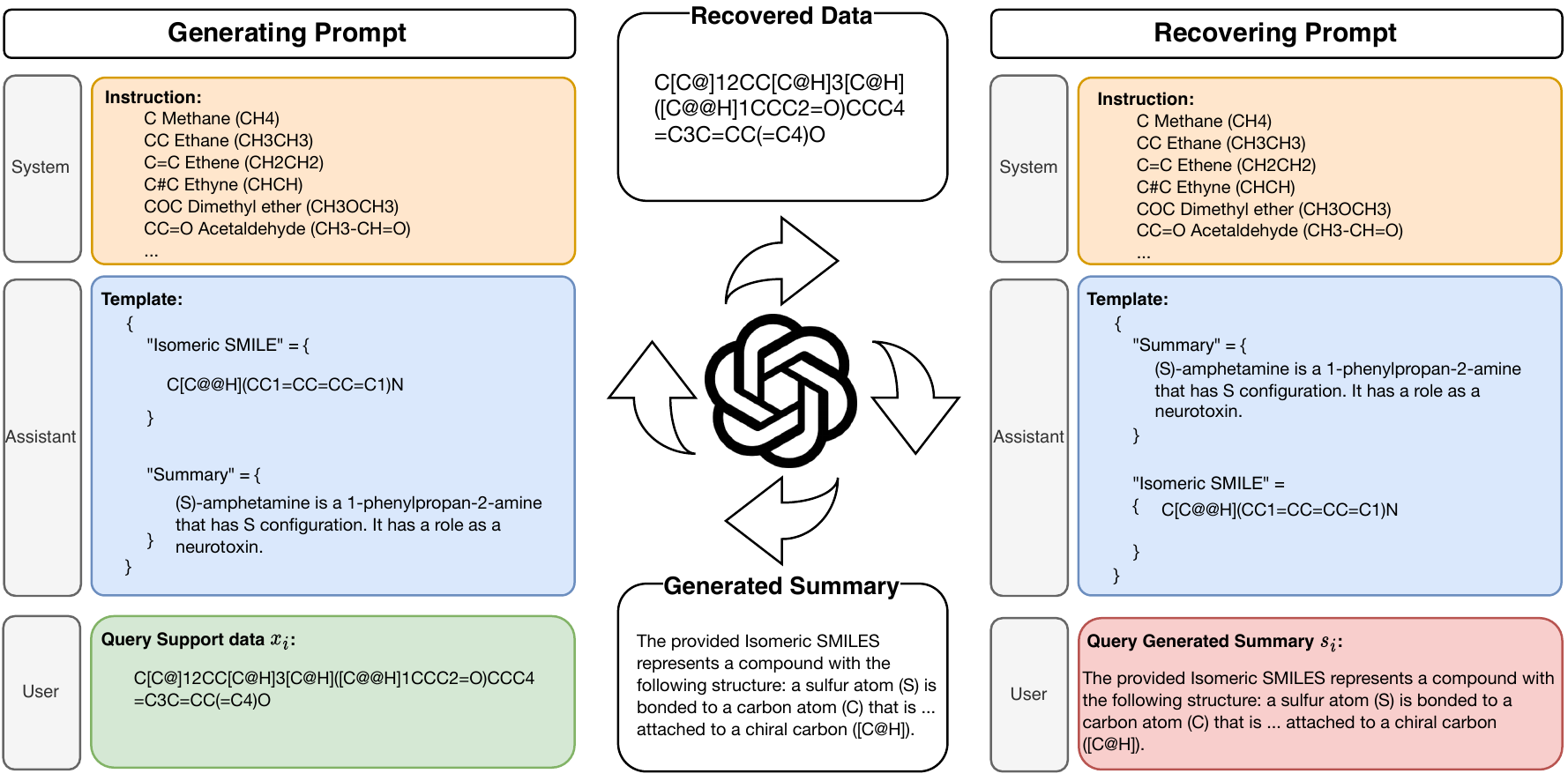}
    \caption{PubMed dataset: A case of the generating-recovering annotation.}
  \label{fig:case2}
\end{figure}

\section{Limitations}

The constraints imposed by token limits restrict the quantity of templates we can employ, posing a significant challenge when conducting experiments within a few-shot framework. Thus, there is a necessary trade-off between the number of shot samples and the length of each sample.

\section{Conclusion}

    This paper introduces a novel approach named 
    GPT self-supervision annotation, 
    which harnesses the one-shot learning capabilities of GPT models to produce concise summaries and alleviate the burden of time and specialized expertise required by human annotators when dealing with complex structured data, such as graphs.
    Our approach consists of two phases: one-shot tuning and generation.
    During the one-shot tuning stage, a support set and a validation set are created from the training data, a template is selected from the support set and used as a prompt to generate a textual summary using GPT models, and the same model is utilized to recover the original data from the generated summary, with alignment scores being calculated for feedback and potential template modification.
    During the generation stage, our approach employs a selected one-shot sample as a template to generate summaries for challenging datasets.
    Both sentence-level (BLEU, ROUGE) and structure-level (STS, BERT) alignment scores between the original and recovered data are assessed, which demonstrates that our approach consistently achieves competitive evaluation scores.
    The results demonstrate the effectiveness of GPT models in data-to-summary annotation tasks.


\newpage
\setcounter{section}{0}
\renewcommand{\thesection}{\Alph{section}}

\section{Appendix} \label{sec:Appendix}
\begin{figure}[h]
    \centering
    \begin{subfigure}{\textwidth}
        \centering
        \includegraphics[width=\textwidth]{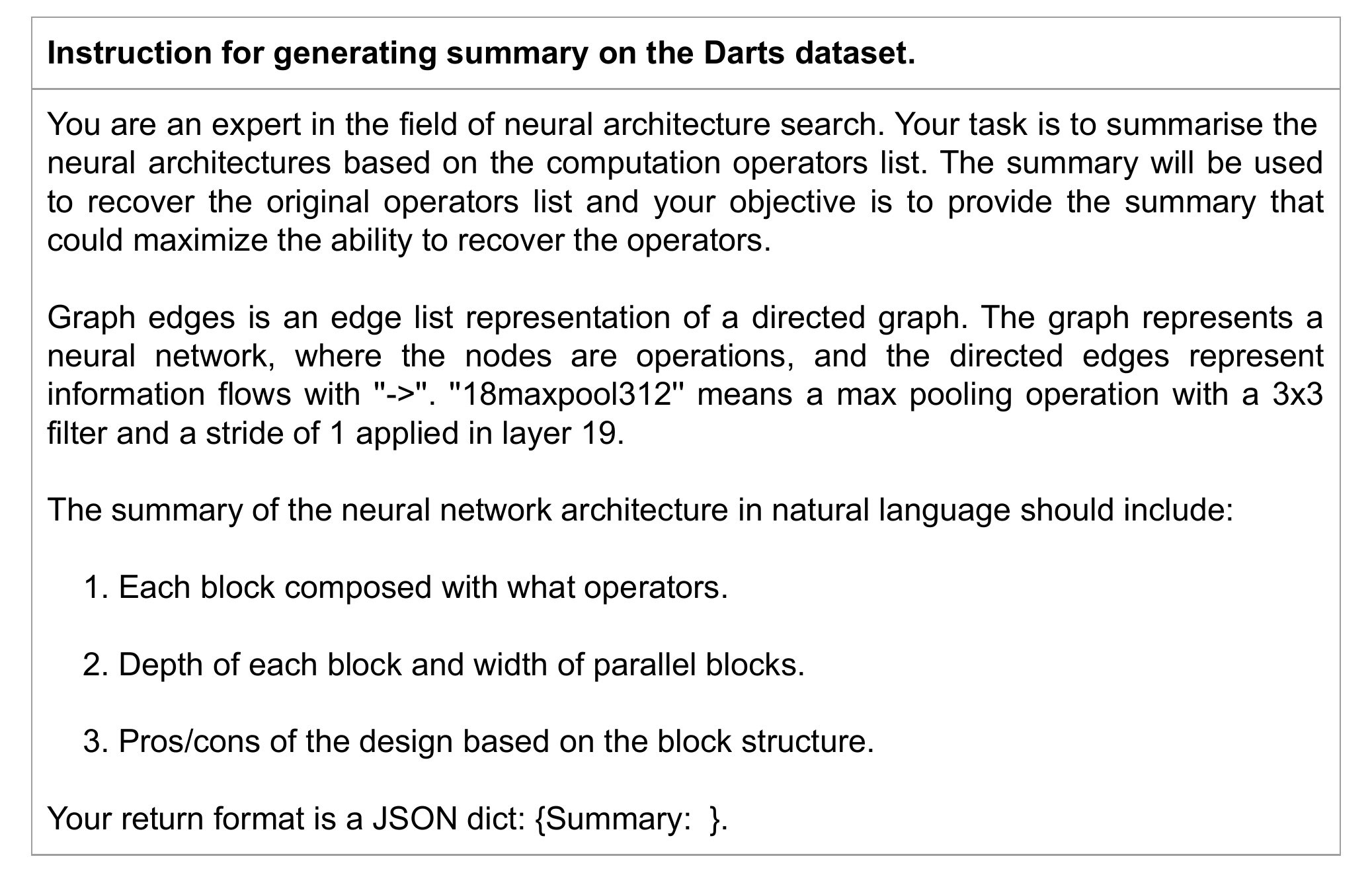}
        \caption{Encoding instruction.}
    \end{subfigure}
    \begin{subfigure}{\textwidth}
        \centering
        \includegraphics[width=\textwidth]{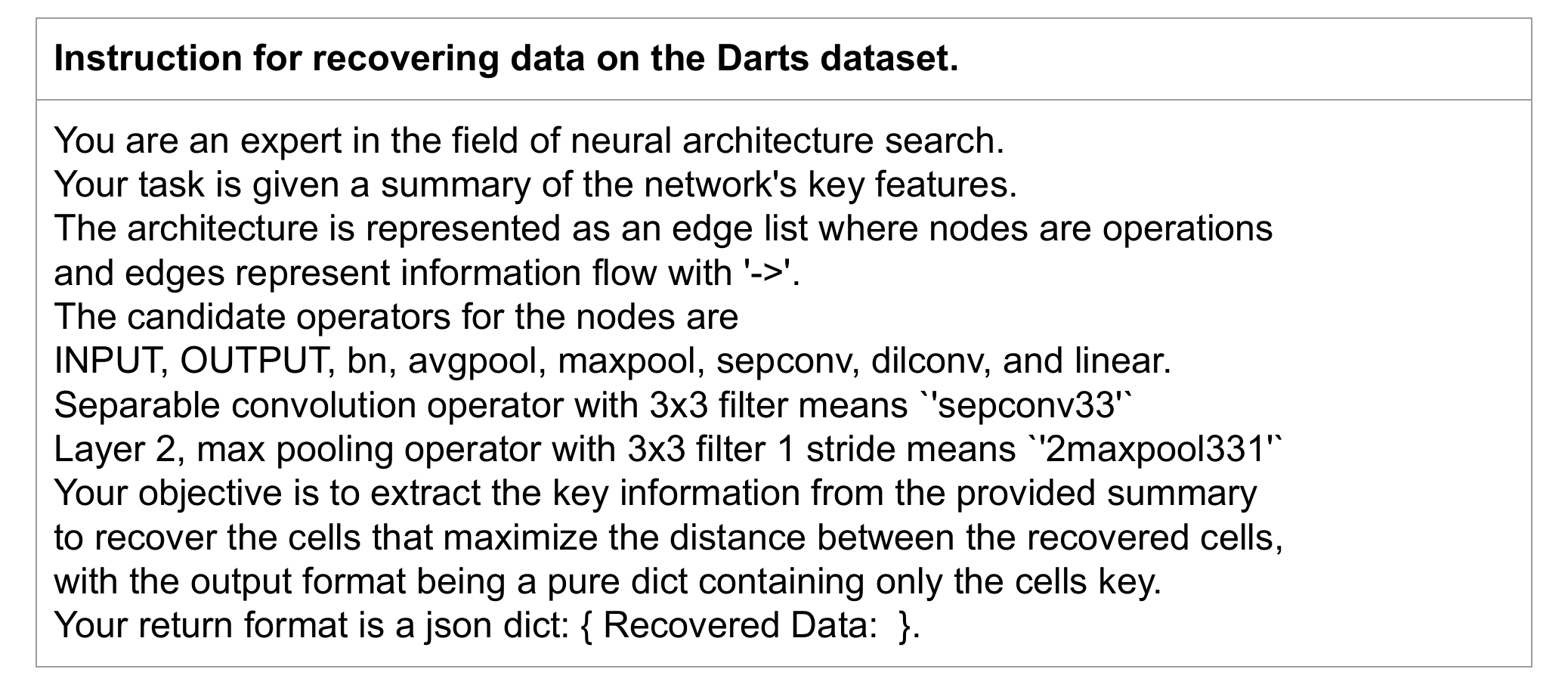}
        \caption{Decoding instruction.}
    \end{subfigure}
    \caption{Instructions for the annotation on the Darts dataset.}
    \label{fig:darts}
\end{figure}

\begin{figure}[h]
    \centering
    \begin{subfigure}{\textwidth}
        \centering
        \includegraphics[width=\textwidth]{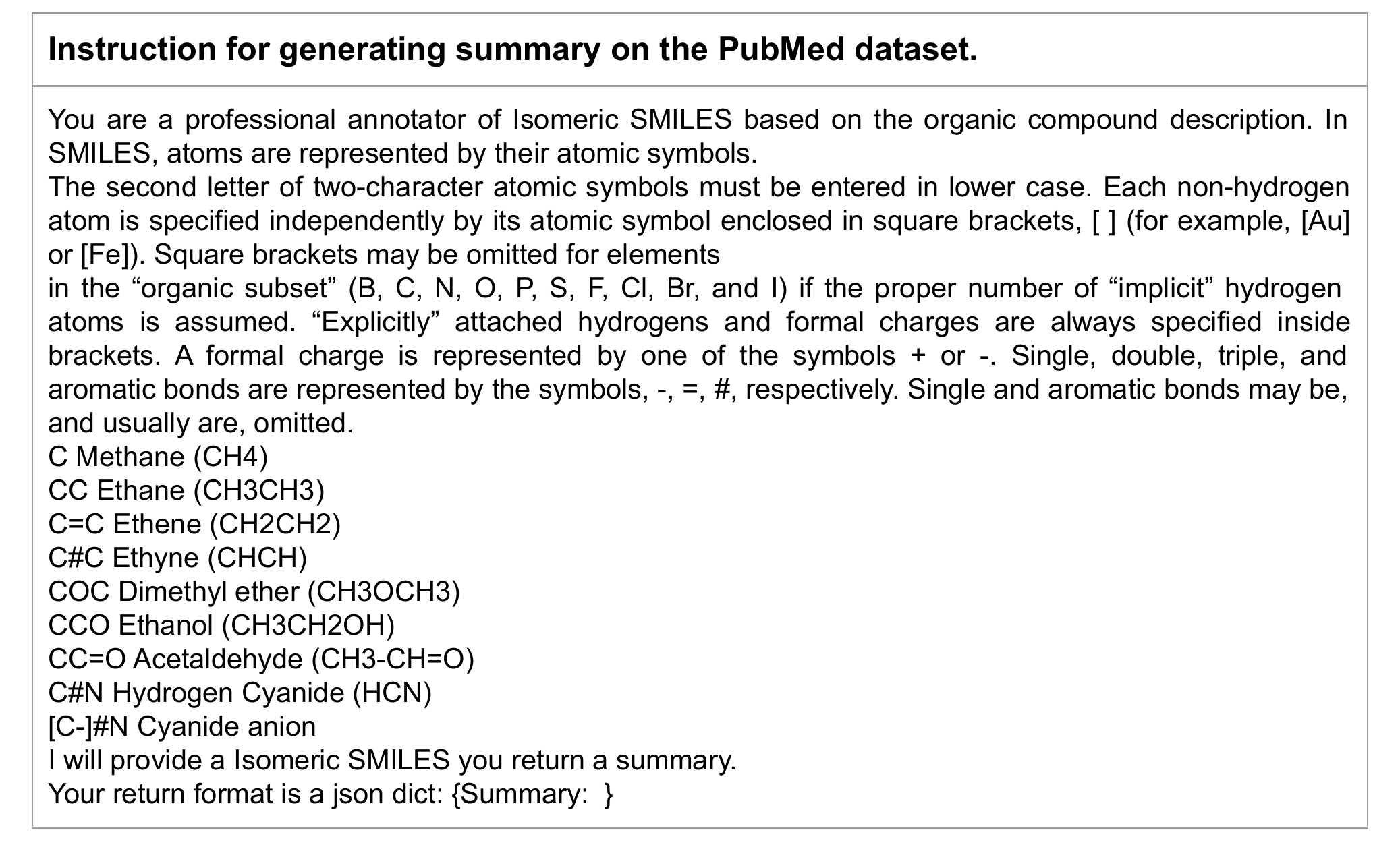}
        \caption{Encoding instruction.}
    \end{subfigure}
    \begin{subfigure}{\textwidth}
        \centering
        \includegraphics[width=\textwidth]{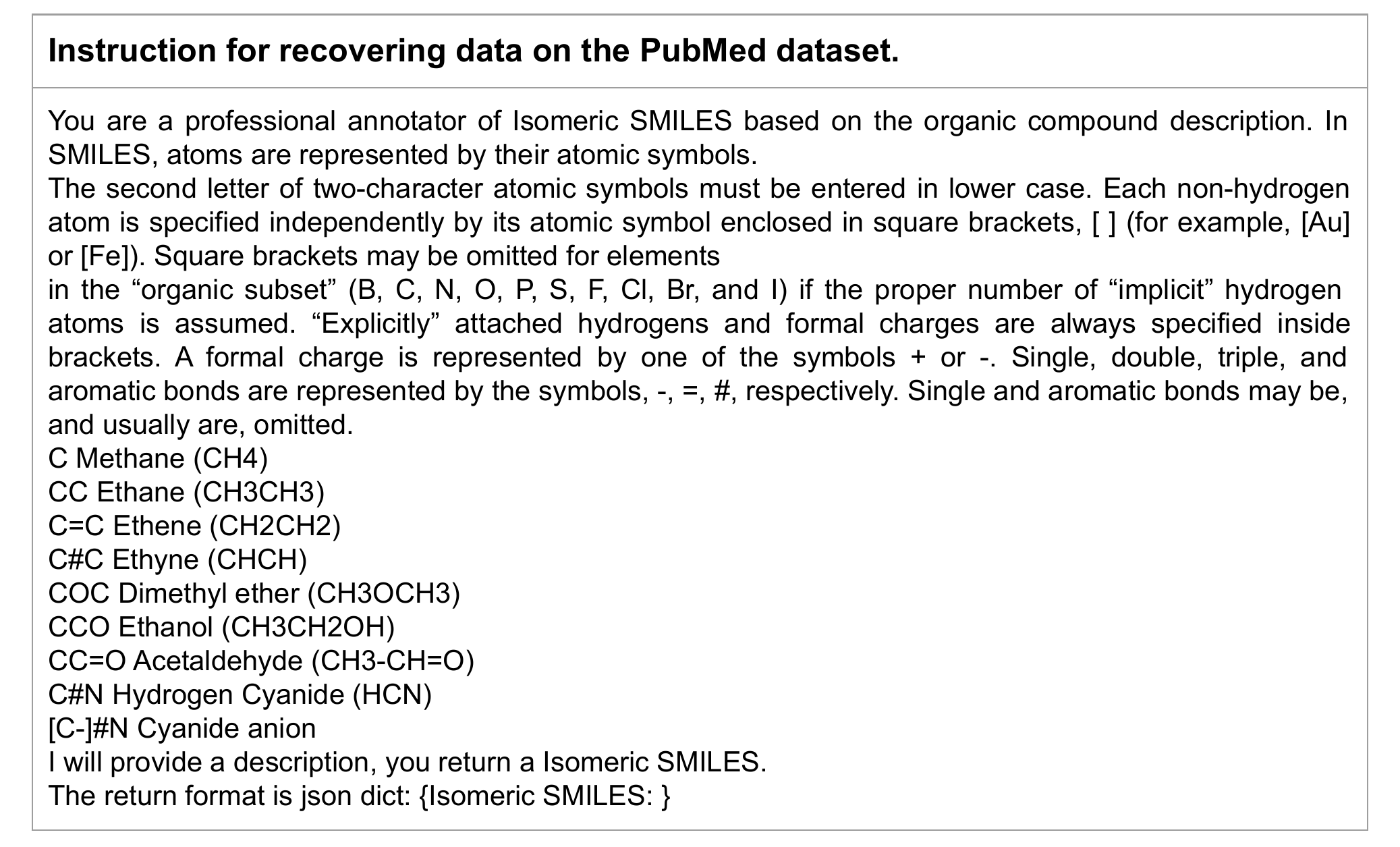}
        \caption{Decoding instruction.}
    \end{subfigure}
    \caption{Instructions for the annotation on the PubMed dataset.}
    \label{fig:pubmed}
\end{figure}

Here we provide the encoding instruction and decoding instruction for generating and recovering on the darts dataset in Figure \ref{fig:darts}.
\newpage
And we also provide the encoding instruction and decoding instruction for generating and recovering on the PubMed dataset in Figure \ref{fig:pubmed}.

\end{document}